\documentclass[runningheads]{llncs}

\usepackage{mypackages}
\pdfoutput=1

\usepackage{graphicx}
\usepackage{amsmath,amssymb} 
\usepackage{color}
\usepackage[width=122mm,left=12mm,paperwidth=146mm,height=193mm,top=12mm,paperheight=217mm]{geometry}
\usepackage{upgreek}

\def\vspaceafterfigure{\vspace{0em}}
\def\vspaceaftertable{\vspace{0em}}
\def\vspaceafteralgo{\vspace{0em}}

\let\llncssubparagraph\subparagraph
\let\subparagraph\paragraph
\usepackage[]{titlesec} 
\let\subparagraph\llncssubparagraph

\titlespacing{\section}{0pt}{*2}{*1}
\titlespacing{\subsection}{0pt}{*2}{*1}
\titlespacing{\subsubsection}{0pt}{*2}{*1}

\def\myalgoname{\mathrm{TMA}}

\def\image{\mathbf{I}}
\def\imageprobe{\image_p}
\def\imagegallery{\image_g}

\def\probeset{\mathcal{P}}
\def\galleryset{\mathcal{G}}

\def\feat{\mathbf{x}}
\def\featprobe{\feat_p}

\def\featgallery{\feat_g}

\def\featpair{(\featprobe, \featgallery; \labelpair)}

\def\featpairstrainingset{\mathcal{X}}

\def\labell{y}
\def\labelpair{\labell_{\pair}}
\def\pair{p,g}
\def\pairi{p^{(i)},g^{(i)}}

\def\pairiter{p^{(t)},g^{(t)}}

\def\lossfun{\mathcal{J}}
\def\lossfunpar{\lossfun_{\kernellowrank,\metriclowrank}}

\def\hinge{\ell}
\def\hingepar{\hinge_{\kernellowrank,\metriclowrank}}
\def\hingepairpar{\hingepar(\pair)}

\def\dissimfun{\delta}
\def\simfun{\sigma}
\def\disssimcomb{S}
\def\disssimcombpar{\disssimcomb_{\kernellowrank,\metriclowrank}}
\def\disssimcombpairpar{\disssimcombpar(\pair)}

\def\learningrate{\eta}

\def\regularizerfun{\Omega}
\def\regularizerfunpar{\regularizerfun_{\kernellowrank,\metriclowrank}}
\def\regularizerfunparlagr{\regularizerfun_{\kernellowranklagr,\metriclowranklagr}}

\def\kernellowrank{\mathbf{K}}

\def\kernellowrankepoch{\kernellowrank^{(s)}}
\def\kernellowrankepochupd{\kernellowrank^{(s+1)}}
\def\kernellowrankpar{\alpha}

\def\kernellowrankapprox{\mathbf{\tilde{K}}}

\def\kernellowrankapproxiter{\kernellowrankapprox^{(t)}}
\def\kernellowrankapproxiterupd{\kernellowrankapprox^{(t+1)}}

\def\kernellowranklagr{\mathbf{U}}

\def\kernellowranklagrepoch{\kernellowranklagr^{(s)}}
\def\kernellowranklagrepochupd{\kernellowranklagr^{(s+1)}}
\def\kernellowranklagrmult{\mathbf{\Lambda}}

\def\kernellowranklagrmultepoch{\kernellowranklagrmult^{(s)}}
\def\kernellowranklagrmultepochupd{\kernellowranklagrmult^{(s+1)}}

\def\metriclowrank{\mathbf{P}}

\def\metriclowrankepoch{\metriclowrank^{(s)}}
\def\metriclowrankepochupd{\metriclowrank^{(s+1)}}
\def\metriclowrankpar{\beta}

\def\metriclowrankapprox{\mathbf{\tilde{P}}}
\def\metriclowrankapproxiter{\metriclowrankapprox^{(t)}}
\def\metriclowrankapproxiterupd{\metriclowrankapprox^{(t+1)}}

\def\metriclowranklagr{\mathbf{V}}

\def\metriclowranklagrepoch{\metriclowranklagr^{(s)}}
\def\metriclowranklagrepochupd{\metriclowranklagr^{(s+1)}}
\def\metriclowranklagrmult{\mathbf{\Psi}}

\def\metriclowranklagrmultepoch{\metriclowranklagrmult^{(s)}}
\def\metriclowranklagrmultepochupd{\metriclowranklagrmult^{(s+1)}}

\def\lagrangian{L}
\def\admmpenalty{\rho}

\def\graph{G}
\def\vertex{V}
\def\edge{E}
\def\adjacencymat{\mathbf{W}}
\def\participationvecelement{h}

\def\participationvec{\mathbf{\participationvecelement}}
\def\participationvecestimate{\hat{\participationvec}}
\def\repdynthreshold{\epsilon}
\def\dominantset{\mathcal{D}}
\def\probegalleryset{\mathcal{H}}

\begin{document}
\pagestyle{headings}
\mainmatter

\title{Temporal Model Adaptation for\\ Person Re-Identification\thanks{The work has been accepted for publication in ECCV 2016. The final publication will be available at Springer. 
		}}

\titlerunning{Temporal Model Adaptation for Person Re-Identification}

\authorrunning{Niki Martinel, Abir Das, Christian Micheloni and Amit K. Roy-Chowdhury}

\author{Niki Martinel\inst{1,}\inst{3} \and Abir Das\inst{2} \and \\Christian Micheloni\inst{1} \and Amit K. Roy-Chowdhury\inst{3}}

\institute{University of Udine, 33100 Udine, Italy\\
	\and
	University of Massatchussets Lowell, 01852 Lowell, MA, USA\\
	\and
	University of California Riverside, 92507 Riverside, CA, USA\\
	}

\maketitle

\begin{abstract}
	Person re-identification is an open and challenging problem in computer vision.
	Majority of the efforts have been spent either to design the best feature representation or to learn the optimal matching metric.
	Most approaches have neglected the problem of adapting the selected features or the learned model over time.
	To address such a problem, we propose a temporal model adaptation scheme with human in the loop.
	We first introduce a similarity-dissimilarity learning method which can be trained in an incremental fashion by means of a stochastic alternating directions methods of multipliers optimization procedure.
	Then, to achieve temporal adaptation with limited human effort, we exploit a graph-based approach to present the user only the most informative probe-gallery matches that should be used to update the model.
	Results on three datasets have shown that our approach performs on par or even better than state-of-the-art approaches while reducing the manual pairwise labeling effort by about $80\%$.

\end{abstract}

\section{Introduction}
\label{sec:intro}
Person re-identification is the problem of matching a person acquired by disjoint cameras at different time instants.
The problem has recently gained increasing attention (see~\cite{Vezzani2014} for a recent survey) due to its open challenges like changes in viewing angle, background clutter, and occlusions. 
To address these issues, existing approaches seek either the best feature representations (\eg,~\cite{Wu2014,Lisanti2014,Martinel2014a}) or propose to learn optimal matching metrics (\eg,~\cite{Xiong2014,Liao2015,Paisitkriangkrai2015}).
While they have obtained reasonable performance on commonly used datasets (\eg,~\cite{Liu,Chen2015,Garcia2015}), we believe that these approaches have not yet considered a fundamental related problem: how to learn from the data being continuously collected in an installed system and adapt existing models to this new data.
This is an important problem to address if re-identification methods have to work on long time-scales.

To illustrate such a problem, let us consider a simplified scenario in which at every time instant a conspicuous amount of visual data is being generated from two cameras.
From each camera we obtain a large set of \textit{probe} and \textit{gallery} persons that have to be matched.
Since this is a task that evolves over time, it is unlikely that the \textit{a-priori} selected features or the learned model return the correct gallery match for every probe at any instant.
In addition, after each of such matches is computed, the information provided by the considered images is discarded.
This results in a loss of valuable information which could have been used to update the model, thus ideally yielding better performance over time.

The above problem could be overcome if the data could be exploited in a continuous learning process in which the model can be updated with every single \textit{probe-gallery} match.
Since we do not know whether a match is correct or not, the model might be updated with the wrong information.
To tackle this issue, manual labeling of each match can be performed, but, doing so with a large corpus of data is clearly impossible. 
However, if the human labor is kept to a minimum, the model can ideally be adapted over time without compromising performance.
Thus, \emph{the main idea of the paper is a person re-identification solution based on an incremental adaptation of the learned model with human in the loop}.

\begin{figure}[!t]
\centering
\includegraphics[width=1\linewidth,height=9.5em]{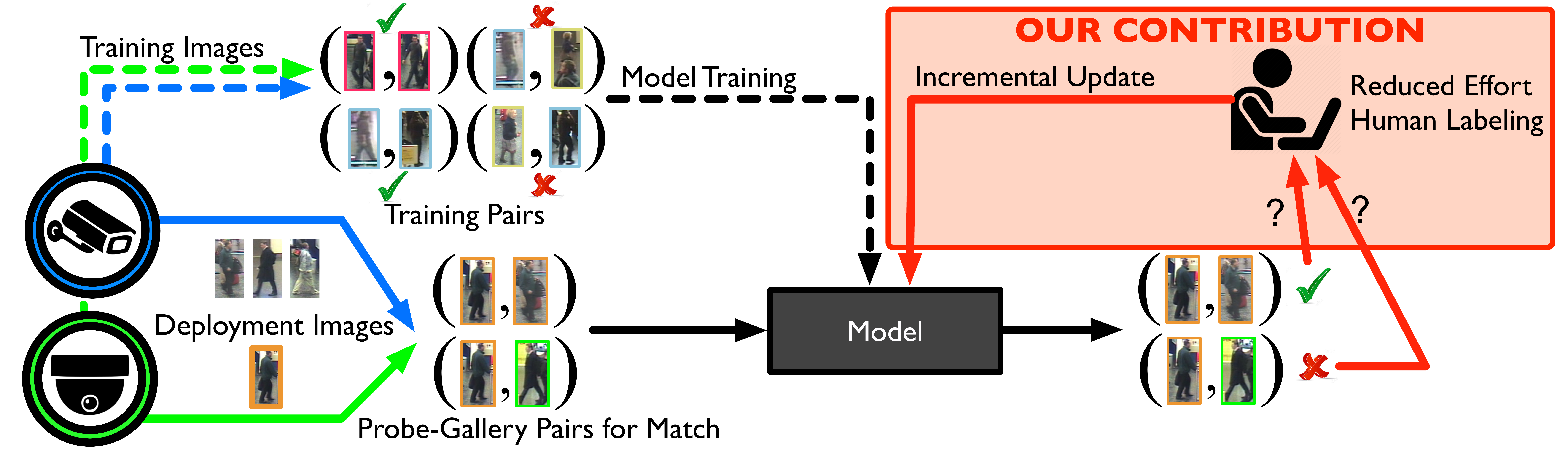}
\caption{Illustration of the re-identification pipeline highlighting our contribution.
Dashed lines indicate the training stage, solid lines the deployment stage.
Existing methods do not consider the information provided by a matched probe\hyp{}gallery pair to update the model.
We propose to use such information to improve the model performance by adapting it to the dynamic environmental variations.
}
\label{fig:contribution}
\vspaceafterfigure
\end{figure}

\noindent\textbf{Contributions:}
As shown in \figurename\ref{fig:contribution}, this work brings in two main contributions:
\begin{enumerate*}[label=(\roman{*})]
\item an incremental learning algorithm that allows the model to be adapted over time, and
\item a method to reduce the human labeling effort required to properly update the model.
\end{enumerate*}
These objectives are achieved as follows.
\begin{enumerate}[label=\roman{*}),topsep=0pt,itemsep=1pt]
\item We propose a low-rank sparse similarity-dissimilarity metric learning method (Section~\ref{sec:sub:metric}) which 
\begin{enumerate}[label=\alph{*})]
\item learns two low-rank projections onto discriminant manifolds providing optimal embeddings for a similarity and a dissimilarity measure;
\item introduces sparsity inducing regularizers that allow identification and exploitation of the most discriminative dimensions for matching; and 
\item is trained in an incremental fashion through a stochastic derivation of the Alternating Directions Methods of Multipliers (ADMM)~\cite{Boyd2010}.
\end{enumerate}
\item We introduce an unsupervised graph-based approach which, for every probe, identifies only the most relevant gallery persons among a large set of available ones (Section~\ref{sec:sub:al}).
Such a set, obtained by exploiting dominant sets clustering~\cite{Pavan2007}, contains the most informative gallery persons which are first provided to the human labeler, then exploited to update the model.
\end{enumerate}

\noindent To substantiate our contributions we have conducted the experiments on three benchmark datasets for person re-identification.
Results demonstrate that
(i) the proposed approach for identifying the most informative gallery persons yields better re-identification performance than using completely labeled data;
(ii) the proposed low-rank sparse similarity-dissimilarity approach trained in an incremental fashion with such informative gallery persons, hence with significantly less manual labor, performs on par or even better than state-of-the-art methods trained on $100\%$ labeled data.
In fact, with only $15\%$ labeled data we improve the previous best rank 1 results by more than $8\%$ on the PRID450S dataset.
These experiments show how re-identification models can be continuously adapted over time with limited human effort and without sacrifice in performance.

%

\section{Relation to Existing Work}
\label{sec:rel_work}
The person re-identification problem has been studied from different perspectives, ranging from partially seen persons~\cite{Zheng2015b} to low resolution images~\cite{Li2015} --also considered in camera networks~\cite{Martinel2016a}, which can eventually be synthesized in the open-world re-identification idea~\cite{Zheng2015a}.
In the following, we focus on metric and active learning methods relevant to our work.

\noindent\textbf{Metric Learning} approaches focus on learning discriminant metrics which aim to yield an optimal matching score/distance between a gallery and a probe image.

Since the early work of~\cite{Xing2002}, many different solutions have been introduced~\cite{Bellet2013}.
In the re-identification field, metric learning approaches have been proposed by relaxing~\cite{Hirzer2012a} or enforcing~\cite{Liao2015a} positive semi-definite (PSD) conditions as well as by considering equivalence constraints~\cite{Kostinger2012,Tao2013,Tao2014}.
While most of the existing methods capture the global structure of the dissimilarity space, local solutions~\cite{Li2013b,Pedagadi2013,Martinel2014,Garcia2016} have been proposed too.
Following the success of both approaches, methods combining them in ensembles~\cite{Paisitkriangkrai2015,Xiong2014,Martinel2015c} have been introduced.

Different solutions yielding similarity measures have also been investigated by proposing to learn listwise~\cite{Chen2015a} and pairwise~\cite{Zheng2012a} similarities as well as mixture of polynomial kernel-based models~\cite{Chen2015}.
Related to these similarity learning models are the deep architectures which have been exploited to tackle the task~\cite{Li2014,Ahmed2015,Zhang2015}.

With respect to all such methods, the closest ones to our approach are~\cite{Liao2015} and~\cite{Liao2015a}.
Specifically, in~\cite{Liao2015}, authors jointly exploit the metric in~\cite{Kostinger2012} and learn a low-rank projection onto a subspace with discriminative Euclidean distance.
The solution is obtained through generalized eigenvalue decomposition.
In~\cite{Liao2015a}, a soft-margin PSD constrained metric with low-rank projections is learned via a proximal gradient method.
Both works exploit a batch optimization approach.

Though sharing the idea of finding discriminative low-rank projections, there are significant differences with our method.
Specifically, we introduce
(i) an incremental learning procedure along with a stochastic ADMM solver which can handle noisy observations of the true data;
(ii) a low-rank similarity\hyp{}dissimilarity metric learning which brings significant performance gain with respect to each of its components;
(iii) additional sparsity regularizers on the low-rank projections that allow self-discovery of the relevant components of the underlying manifold.

\noindent\textbf{Active Learning:}
In an effort to bypass tedious labeling of training data there has been recent interest in ``active learning''~\cite{Settles2012} to intelligently select unlabeled examples for the experts to label in an interactive manner.

This can be achieved by choosing one sample at a time by maximizing the value of information~\cite{Joshi2012}, reducing the expected error~\cite{Aodha2014}, or minimizing the resultant entropy of the system~\cite{Biswas2013}.
More recently, works selecting batches of unlabeled data by exploiting classifier feedback to maximize informativeness and sample diversity~\cite{Chakraborty2011,Elhamifar2013} were proposed.
Specific application areas in computer vision include, but are not limited to, tracking~\cite{Vondrick2011}, scene classification~\cite{Joshi2012,Vijayanarasimhan2011}, semantic segmentation~\cite{Vezhnevets2012}, video annotation~\cite{Karasev2014} and activity recognition~\cite{Hasan2015}.

Active learning has been a relatively unexplored area in person re-identification.
Including the human in the loop has been investigated in~\cite{Liu,Wang2014c,Das2015}.
These methods focused on post-ranking solutions and exploit human labor to refine the initial results by relying on full~\cite{Liu} or partial~\cite{Wang2014c} image selection.
In~\cite{Das2015}, authors introduce an active learning strategy that exploits mid level attributes to train a set of attribute predictors aiding active selection of images.

Different from such approaches, in our proposed method human labor is not required to improve the post-rank visual search, but to reliably update the learned model over time. We do not rely on additional attribute predictors which require a proper training that calls for a large number of annotated attributes.
Thus bypassing the need for attribute annotation, we reduce both the computational complexity as well as the additional manual effort.
We introduce a graph-based solution that exploits the information provided by a single probe-gallery match as well as the information shared between all the persons in the entire gallery.
With this, a small set of highly informative probe-gallery pairs is delivered to the human, whose effort is thus limited.

\section{Temporal Model Adaptation for Re-Identification}
\label{sec:approach}
\begin{figure}[!t]
\centering
\includegraphics[width=1\linewidth,height=11.5em]{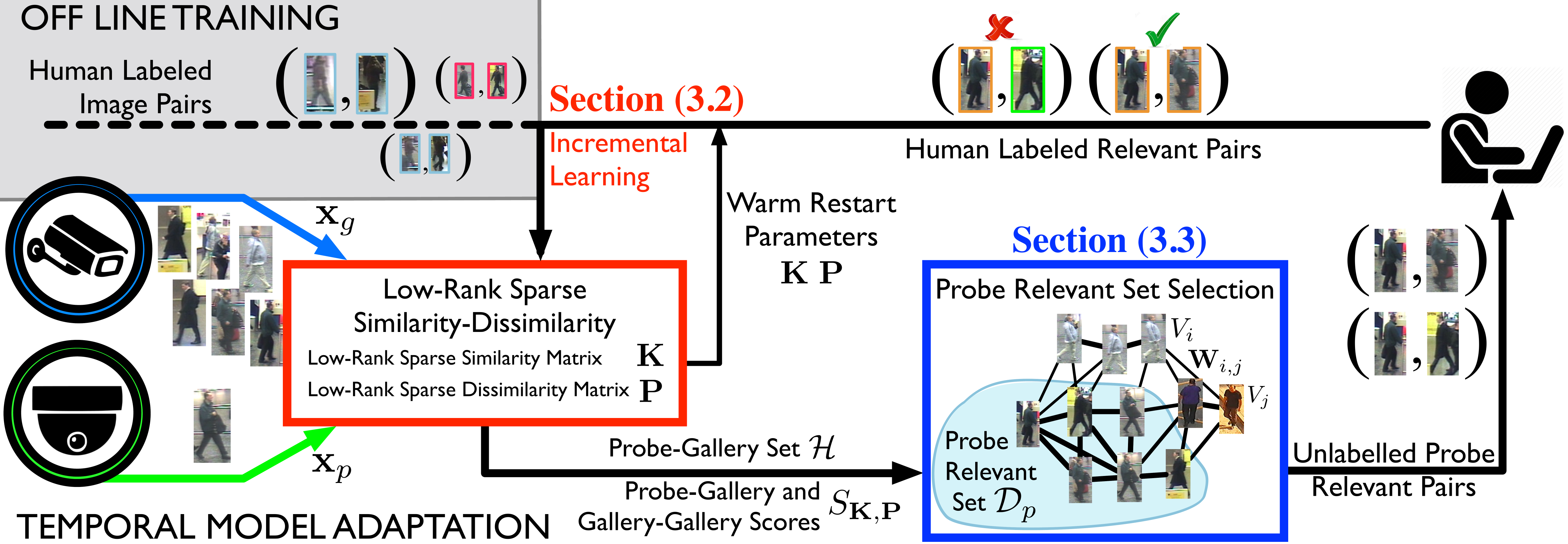}
\caption{Proposed temporal model adaptation scheme.
An off-line procedure exploits labeled image pairs to train the initial similarity-dissimilarity model.
As new unlabeled pairs are obtained, a score for each of those is obtained using the learned model.
These are later used to identify a relevant set of gallery persons for each probe.
Such a set, containing the most informative samples, is exploited to construct the relevant pairs which are first provided to the human annotator, then considered to update the model.}
\label{fig:system}
\vspaceafterfigure
\end{figure}
An overview of the proposed solution is illustrated in \figurename\ref{fig:system}.
Specifically, to achieve model adaptation over time, we first introduce a similarity\hyp{}dissimilarity metric learning approach which can be trained in an incremental fashion (Section~\ref{sec:sub:metric}).
Then, to limit the human labeling effort required to properly update the model, we propose an unsupervised graph-based approach that identifies only the most informative probe-gallery samples (Section~\ref{sec:sub:al}).

\subsection{Preliminaries}
\label{sec:sub:preliminaries}
Let $\probeset = \{\imageprobe\}_{p=1}^{|\probeset|}$ and $\galleryset = \{\imagegallery\}_{g=1}^{|\galleryset|}$ be the set of probe and gallery images acquired by two disjoint cameras.
Let $\featprobe \in \mathbb{R}^{d}$ and $\featgallery \in \mathbb{R}^{d}$ be the feature representations of $\imageprobe$ and $\imagegallery$ of two persons $p$ and $g$.
Let $\featpairstrainingset = \{\featpair^{(i)}\}_{i=1}^{n}$ denote the training set of $n = |\probeset|\times|\galleryset|$ probe-gallery pairs where $\labelpair \in \{-1, +1\}$ indicates whether $p$ and $g$ are the same person ($+1$) or not ($-1$).
Finally, let an \textit{iteration} be a parameter update computed by visiting a single sample and let an \textit{epoch} denote a complete cycle on the training set.

\subsection{Low-Rank Sparse Similarity-Dissimilarity Learning}
\label{sec:sub:metric}
\noindent\textbf{Objective:}
The image feature representations $\feat$ might be very high-dimensional and contain non-discriminative components.
Hence, learning a metric in such a feature space might yield to non-optimal generalization performance.
To overcome such a problem we propose to learn a low-rank metric which self-determines the discriminative dimensions of the underlying manifold.

Towards such an objective, inspired by the success of similarity learning on image retrieval tasks~\cite{Chechik2010,Guo2013,Xia2014}, we propose to learn a similarity function
\begin{equation}
\label{eq:sim}
\simfun_{\kernellowrank}(\featprobe, \featgallery) = \featprobe^T \kernellowrank^T \kernellowrank \featgallery
\end{equation}
parameterized by the low-rank projection matrix $\kernellowrank \in \mathbb{R}^{r\times d}$, with $r \ll d$.
This provides an embedding in which the dot product between the projected feature vectors is ``large'' if $p$ and $g$ are the same person, ``small'' otherwise.
The similarity function is then coupled with the output of a metric learning solution that aims to find a matrix $\metriclowrank \in \mathbb{R}^{r\times d}$ that projects the high-dimensional vectors to a low-dimensional manifold with a discriminative Euclidean dissimilarity
\begin{equation}
\label{eq:dissim}
\dissimfun_{\metriclowrank}(\featprobe, \featgallery) = \| \metriclowrank\featprobe - \metriclowrank\featgallery \|_{2}^{2} = (\featprobe-\featgallery)^T \metriclowrank^T \metriclowrank (\featprobe-\featgallery)
\end{equation}
which is ``small'' if $p$ and $g$ are the same person, ``larger'' otherwise.
This results in the score function
\begin{equation}
\label{eq:dissim_sim}
\disssimcombpairpar = 
\labelpair ( \underbrace{\simfun_{\kernellowrank}(\featprobe, \featgallery)}_{ \uparrow \textrm{\scriptsize for } p=g,\ \downarrow \textrm{\scriptsize for } p\neq g} - \underbrace{(1/2) \dissimfun_{\metriclowrank}(\featprobe, \featgallery)}_{\downarrow \textrm{\scriptsize for } p=g,\ \uparrow \textrm{\scriptsize for } p\neq g} )
\end{equation}
which included in a margin hinge loss yields
\begin{equation}
\label{eq:hinge}
\hingepairpar = \max\left(0, 1 - \disssimcombpairpar \right).
\end{equation}
Notice that zero loss is achieved if $\disssimcombpairpar\geq1$,~\ie, when the difference between $\simfun_{\kernellowrank}$ and $\frac{1}{2} \dissimfun_{\metriclowrank}$ is either greater than or equal to $1$ for positive pairs or less than or equal to $-1$ for negative ones.
In other cases a linear penalty is paid.


Obtaining the low-rank projections through eq.(\ref{eq:hinge}) with fixed $r$ implies that such a value should be carefully selected before the learning process begins.
To overcome such a problem, we impose additional constraints on the low-rank projection matrices.
In particular, the $\ell_{2,1}$ norm has shown to perform robust feature selection through the induced group sparsity~\cite{Nie2010,Bach2011,Cao2014,Zhang2015a}.
Motivated by such findings, we can set $r=d$, then leverage on an $\ell_{2,1}$ norm regularizer to drive the rows of $\metriclowrank$ and $\kernellowrank$ to decay to zero.
This corresponds to rejecting non discriminative dimensions of the underlying manifold.

Let $\regularizerfunpar =  \kernellowrankpar\|\kernellowrank\|_{2,1} + \metriclowrankpar\|\metriclowrank\|_{2,1}$ be the cost associated with the low-rank projection matrix regularizers where $\kernellowrankpar$ and $\metriclowrankpar$ are the corresponding trade-off parameters controlling the regularization strength.
Then, considering that we want to optimize the empirical risk over $\featpairstrainingset$, we can write our objective as
\begin{equation}
\label{eq:objective}
\argmin_{\kernellowrank,\metriclowrank} \lossfunpar + \regularizerfunpar \qquad \mbox{where}  \qquad \lossfunpar = \frac{1}{n} \sum_{i=1}^{n} \hingepar\left(\pairi \right)
\end{equation}
and $p^{(i)}$ and $g^{(i)}$ denote the identities of persons $p$ and $g$ in the $i$-th pair of $\featpairstrainingset$.

\noindent\textbf{Incremental Learning:}
The objective function in eq.(\ref{eq:objective}) is a sum of two functions which are both convex but non-smooth.
A solution to such kind of a problem that allows us to perform incremental updates can be obtained using the ADMM optimization algorithm~\cite{Boyd2010}. 

ADMM solves optimization problems defined by means of the corresponding augmented Lagrangian.
By introducing two additional constraints $\kernellowrank-\kernellowranklagr=\mathbf{0}$ and $\metriclowrank-\metriclowranklagr=\mathbf{0}$ we can define the augmented Lagrangian for eq.(\ref{eq:objective}) as
\begin{align}
\label{eq:admm_lagrangian_problem}
\lagrangian_{\kernellowrank, \metriclowrank, \kernellowranklagr, \metriclowranklagr, \kernellowranklagrmult, \metriclowranklagrmult} & = \lossfunpar + \regularizerfunparlagr + \langle \kernellowranklagrmult, \kernellowrank - \kernellowranklagr \rangle + \langle \metriclowranklagrmult,  \metriclowrank - \metriclowranklagr \rangle 
\\ \nonumber
& \qquad + \frac{\admmpenalty}{2} \left( \norm{\kernellowrank-\kernellowranklagr}^{2}_{F} + \norm{\metriclowrank-\metriclowranklagr}^{2}_{F}\right)
\end{align}
where $\kernellowranklagrmult \in \mathbb{R}^{r\times d}$ and $\metriclowranklagrmult \in \mathbb{R}^{r\times d}$ are two Lagrangian multipliers, $\langle\cdot,\cdot\rangle$ denote the inner product, $\norm{\cdot}_{F}$ is the Frobenius norm and, $\admmpenalty > 0$ is a penalty parameter.

To solve the optimization problem, at each epoch $s$, ADMM alternatively minimizes $\lagrangian$ with respect to a single parameter, $\kernellowrank$, $\metriclowrank$, $\kernellowranklagr$, $\metriclowranklagr$, $\kernellowranklagrmult$ or $\metriclowranklagrmult$, keeping others fixed.
The result of each minimization gives the updated parameter.

Standard deterministic ADMM implicitly assumes true data values are available, hence overlooking the existence of noise~\cite{Ouyang2013}.
Noticing that only $\kernellowrank$ and $\metriclowrank$ depend on the data samples, we define the corresponding update rules using the scalable stochastic ADMM approach~\cite{Zhao2015,Johnson2013} which can handle such an issue.

\noindent\uline{\textbf{Update $\kernellowrank$ and $\metriclowrank$}:}
Let $\pder{}{\kernellowrank}\lossfun_{\kernellowrank,\metriclowrank} = \frac{1}{n}\sum_{i=1}^{n} \pder{}{\kernellowrank}\hinge_{\kernellowrank,\metriclowrank}(\pairi)$ and $\pder{}{\metriclowrank}\lossfun_{\kernellowrank,\metriclowrank} = \frac{1}{n}\sum_{i=1}^{n} \pder{}{\metriclowrank}\hinge_{\kernellowrank,\metriclowrank}(\pairi)$ denote the subgradients components of eq.(\ref{eq:hinge}) computed for all samples with respect $\kernellowrank$ and $\metriclowrank$, respectively.
Then, at each iteration $t$,~\ie, for the $t$-th random sample, we compute
\begin{align}
\nonumber
\kernellowrankapproxiterupd & = \kernellowrankapproxiter - \learningrate \Biggl(  \pder{}{\kernellowrankapproxiter}\hinge_{\kernellowrankapproxiter,\metriclowrankapproxiter}(\pairiter) - \pder{}{\kernellowrankepoch}\hinge_{\kernellowrankepoch,\metriclowrankepoch}(\pairiter) 
\\[-1em]
& \qquad \qquad \qquad + \pder{}{\kernellowrankepoch}\lossfun_{\kernellowrankepoch,\metriclowrankepoch}   + \admmpenalty \left(\kernellowrankapproxiter - \kernellowranklagrepoch + \kernellowranklagrmultepoch/\admmpenalty\right)\Biggr)
\label{eq:kernel_iter_upd}
\\ 
\nonumber
\metriclowrankapproxiterupd & = \metriclowrankapproxiter - \learningrate \Biggl( \pder{}{\metriclowrankapproxiter}\hinge_{\kernellowrankapproxiterupd,\metriclowrankapproxiter}(\pairiter) - \pder{}{\metriclowrankepoch}\hinge_{\kernellowrankepoch,\metriclowrankepoch}(\pairiter)
\\[-1em]
& \qquad \qquad \qquad +  \pder{}{\metriclowrankepoch}\lossfun_{\kernellowrankepoch,\metriclowrankepoch}  + \admmpenalty \left(\metriclowrankapproxiter - \metriclowranklagrepoch + \metriclowranklagrmultepoch/\admmpenalty\right)\Biggr)
\label{eq:metric_iter_upd}
\end{align}
where $\learningrate$ is the step size
and $\kernellowrankapproxiter$ and $\metriclowrankapproxiter$ denote the parameters for a specific iteration $t$, while $\kernellowrankepoch$ and $\metriclowrankepoch$ represent the parameters obtained for epoch $s$.
Once $T$ iterations are completed, the two low-rank matrices are updated as
\begin{equation}
\kernellowrankepochupd = \frac{1}{T}\sum_{t=1}^{T}\kernellowrankapproxiter
\qquad \quad
\metriclowrankepochupd = \frac{1}{T}\sum_{t=1}^{T}\metriclowrankapproxiter
\label{eq:admm_kernel_metric_update}
\end{equation}

\noindent\uline{\textbf{Update $\kernellowranklagr$ and $\metriclowranklagr$}:}
To derive the updates for the two regularizers, we first compute the partial derivatives of eq.(\ref{eq:admm_lagrangian_problem}) with respect to $\kernellowranklagr$ and $\metriclowranklagr$ while keeping other parameters fixed.
Then, solving for a stationary point yields
\begin{align}
\kernellowranklagrepochupd & =
\left( \kernellowrankepochupd_{i,:}+ \kernellowranklagrmultepoch_{i,:}/\admmpenalty \right)\max\Bigl(0, 1- \kernellowrankpar / \left( \admmpenalty \norm{ \kernellowrankepochupd_{i,:}+\kernellowranklagrmultepoch_{i,:}/\admmpenalty}_2 \right) \Bigr)
\label{eq:admm_kernel_lagr}
\\
\metriclowranklagrepochupd & =
\left( \metriclowrankepochupd_{i,:}+ \metriclowranklagrmultepoch_{i,:}/\admmpenalty \right)\max \Bigl(0, 1- \metriclowrankpar / \left(\admmpenalty \norm{\metriclowrankepochupd_{i,:}+\metriclowranklagrmultepoch_{i,:}/\admmpenalty}_2 \right) \Bigr)
\label{eq:admm_metric_lagr}
\end{align}
whose closed form solutions have been obtained using the group soft-thresholding technique~\cite{Bach2011} and $i=1,\cdots,r$ denotes the $i$-th row of a parameter matrix.

\noindent\uline{\textbf{Update $\kernellowranklagrmult$ and $\metriclowranklagrmult$}:}
Results from eq.(\ref{eq:admm_kernel_metric_update}) and eq.(\ref{eq:admm_kernel_lagr}-\ref{eq:admm_metric_lagr}) can be finally used to update the duals for the Lagrangian multipliers as 
\begin{align}
\kernellowranklagrmultepochupd & = \kernellowranklagrmultepoch + \admmpenalty ( \kernellowrankepochupd- \kernellowranklagrepochupd )
\label{eq:admm_kernellagrmult}
\\
\metriclowranklagrmultepochupd & = \metriclowranklagrmultepoch + \admmpenalty ( \metriclowrankepochupd- \metriclowranklagrepochupd 
)
\label{eq:admm_metriclagrmult}
\end{align}

To conclude, after $S$ epochs have been performed, the optimal estimates for the two low-rank projection matrices are given by $\kernellowrank^{(S)}$ and $\metriclowrank^{(S)}$.



\subsection{Model Adaptation with Reduced Human Effort}
\label{sec:sub:al}
In the previous section we have presented a similarity-dissimilarity learning model which can be trained in an incremental fashion.
To achieve model adaptation over time, we propose to perform incremental steps to minimize eq.(\ref{eq:admm_lagrangian_problem}) with new image pairs that are progressively acquired as time passes.
This requires human labeling of such pairs.
To limit such a manual effort and improve model generalization, we aim to select only a small set of informative gallery persons to update the model.
These are persons for which the positive/negative association with the probe is very uncertain.
Given a probe, such gallery persons form its \textit{probe relevant set}.

\noindent\textbf{Probe Relevant Set Selection:}
Let $\probegalleryset = \{ \featprobe, \featgallery \ | \ g=1,\ldots,|\galleryset| \}$ denote the probe-gallery set for probe $p$.
We represent such a set as an undirected graph with no loops.
More precisely, let $\graph = (\vertex, \edge, \adjacencymat)$ denote a graph where $\vertex = \{p, g | g=1, \ldots, |\galleryset|\}$ is the set of vertices, $\edge \subseteq \vertex \times \vertex$ is the set of edges and $\adjacencymat \in \mathbb{R}^{|\vertex|\times|\vertex|}_{+}$ denotes the adjacency symmetric matrix of positive edge weights such that, for any two vertices $i$ and $j $, $\adjacencymat_{i,j} = f(\disssimcombpar(i,j))$ if $i\neq j$, $\adjacencymat_{i,j} = 0$, otherwise.
$f(\cdot)$ is the Platt function~\cite{Platt1999} used to ensure a positive edge weight.

To obtain the probe relevant set, we aim to cluster $G$ in such a way that
(i) a cluster contains the probe and gallery persons which are similar to each other, and
(ii) all persons outside a cluster should be dissimilar to the ones inside.
To achieve such an objective, we exploit the dominant sets clustering technique~\cite{Pavan2007}.

Dominant set clustering partitions a graph into dominant sets on the basis of the coherency between vertices as measured by the edge weights.
A dominant set is a subset of the graph nodes having high internal and low external coherency.

To obtain such partitions, the dominant sets approach is based on the participation vector $\participationvec$.
It expresses the probability of participation of the corresponding person in the cluster.
More precisely, the objective is
\begin{equation}
\label{eq:ds_objective}
\participationvecestimate = \argmax_{\participationvec} \participationvec^T \adjacencymat \participationvec
\qquad \qquad \mbox{s.t.} \quad \participationvec \in \mathcal{S}
\end{equation}
where $\mathcal{S}$ is the standard simplex of $\mathbb{R}^{|\vertex|}$.

Let the participation vector be initialized to a uniform distribution,~\ie,$\participationvecelement_i = 1/|\vertex|$, for $i=1,\ldots,|\vertex|$
\footnote{Effect of this initialization is checked by adding random noise to each element of $\participationvec$. Results show that in 96\% of the cases the output cluster is the same.}.
Then, as shown in~\cite{Pavan2007}, a solution to the optimization problem can be obtained by an iterative procedure that, at each iteration $k$, updates the participation vector as
\begin{equation}
\label{eq:replicator_dynamics}
\participationvecelement_i^{(k+1)} = \participationvecelement_i^{(k)} 
\frac{(\adjacencymat\participationvec^{(k)})_i}{(\participationvec^{(k)})^T \adjacencymat \participationvec^{(k)}} \qquad \qquad \mbox{for} \ i = 1, \ldots, |\vertex|
\end{equation}

The iterative updates are applied until the objective function difference between two consecutive iterations is higher than a predefined threshold $\repdynthreshold$.
When such a condition is not satisfied a local optima is obtained and the non-zero entries in the participation vector $\participationvecestimate$ specify the relevant nodes included in the dominant set.
Notice that the dominant sets clustering can be easily extended to cluster a graph in multiple dominant sets.
This is obtained by removing the person identities included in the current dominant set from $\probegalleryset$, creating the new graph structure and then repeating the process.
In our approach such a procedure is applied until the dominant set containing the probe person $p$ is found.
This is the probe relevant set for person $p$ and is denoted as $\dominantset_{p} = \{ i \ | \ i \neq p \wedge \participationvecelement_i > 0 \}$.

\noindent\textbf{Incremental Model Update:}
Armed with the probe relevant set, we can now achieve temporal model adaptation by performing the incremental learning steps described in sec.~\ref{sec:sub:metric}.
Towards this objective, we first ask the human annotator to label only the probe relevant pairs in $\{(\imageprobe, \imagegallery) \ | \ g \in \dominantset_p \}$.
Then, using the current parameters $\kernellowrank$ and $\metriclowrank$ as a ``warm-restart'', we exploit the newly labeled samples to run $\hat{S}$ epochs, each providing $\hat{T}$ incremental iterations.
When such a process is completed the updated model parameters $\kernellowrank$ and $\metriclowrank$ are obtained.

\subsection{Discussion}
Through the preceding sections we have introduced two main contributions that allow us to obtain model adaptation over time.
Specifically, the goal has been achieved (i) by proposing a stochastic similarity-dissimilarity metric learning procedure that can be incrementally updated and (ii) by introducing a graph-based approach that allows to identify the most informative pairs that should be labeled by the human.
All the steps are summarized in Algorithm~\ref{algo:all}.
\begin{algorithm}[t]
\scriptsize
 \SetKwInput{KwSet}{Set}
 \SetKwRepeat{KwRepeat}{Repeat}{Until}
 \SetKwFor{KwIterate}{Iterate}{}{}
 \SetKw{KwOfflinePart}{\uline{Off-Line Training\hspace{28em}}}  \SetKw{KwOnlinePart}{\uline{Temporal Model Adaptation\hspace{23em}}}
 \SetKwInput{KwInit}{Initialize}
 \KwOfflinePart{}
 
\KwIn{$\featpairstrainingset$, $\learningrate > 0$, $\admmpenalty > 0$, $T>0$, $S>0$}
\KwOut{Discriminative low rank projection matrices $\kernellowrank$ and $\metriclowrank$}
\KwInit{$\kernellowrank^{(1)}$ and $\metriclowrank^{(1)}$ to random, $\kernellowranklagrmult^{(1)}$ and $\metriclowranklagrmult^{(1)}$ to $\mathbf{0}$}
 \KwSet{$\kernellowranklagr^{(1)}=\kernellowrank^{(1)}$,$\metriclowranklagr^{(1)}=\metriclowrank^{(1)}$}
 \KwIterate{for $s=1, \ldots, S$}
 {
 \textbf{1.} Consider all the $n$ training samples to pre-compute the average hinge loss subgradients with respect to $\kernellowrankepoch$ and $\metriclowrankepoch$ 
 \\
 \textbf{2.} Set $\kernellowrankapproxiter=\kernellowrankepoch$, $\metriclowrankapproxiter=\metriclowrankepoch$, then run $T$ iterations and update $\kernellowrankapproxiter$ and $\metriclowrankapproxiter$ as in eq.(\ref{eq:kernel_iter_upd}) and eq.(\ref{eq:metric_iter_upd})
 \\
 \textbf{3.} Average over the $T$ updates as in eq.(\ref{eq:admm_kernel_metric_update}) to obtain $\kernellowrankepochupd$ and $\metriclowrankepochupd$
 \\
 \textbf{4.} Update the constraints $\kernellowranklagrepoch$ and $\metriclowranklagrepoch$ using eq.(\ref{eq:admm_kernel_lagr}) and eq.(\ref{eq:admm_metric_lagr}) 
 \\
 \textbf{5.} Compute the dual updates for the Lagrangian multipliers as in eq.(\ref{eq:admm_kernellagrmult}) and eq.(\ref{eq:admm_metriclagrmult})
 }
 \textbf{6.} Obtain the optimal estimates $\kernellowrank=\kernellowrank^{(S)}$ and $\metriclowrank=\metriclowrank^{(S)}$
\\
 \vspace{0.5em}
 \KwOnlinePart{}
 
 \KwIn{$\kernellowrank$, $\metriclowrank$, $\mathcal{H}$, $\learningrate > 0$, $\admmpenalty > 0$, $\hat{T}>0$, $\hat{S}>0$, $\epsilon>0$}
 \KwOut{Updated discriminative low rank projection matrices $\kernellowrank$ and $\metriclowrank$}
 \textbf{1.} Compute the scores for each possible probe-gallery pair via $\disssimcombpar$ to obtain $\adjacencymat$
 \\
 \textbf{2.}Solve the problem in eq.(\ref{eq:ds_objective}) --using eq.(\ref{eq:replicator_dynamics})-- to obtain the probe relevant set $\dominantset_{p}$
 \\
 \textbf{3.} Form the set of probe relevant pairs
 \\
 \textbf{4.} Update $\kernellowrank$ and $\metriclowrank$ by performing \textit{off-line training} steps \textbf{1-6} with the probe relevant pairs, $S=\hat{S}$ and $T=\hat{T}$
 \\
 \caption{Temporal Model Adaptation for Person Re-Identification}
 \label{algo:all}
\end{algorithm}
\vspaceafteralgo


MLAPG~\cite{Liao2015a} and XQDA~\cite{Liao2015}, which learn a discriminant subspace as well as a distance function in the learned subspace, are close to the proposed approach.
However, both of them do not update the model over time.
In addition, our solution differs in the stochastic ADMM optimization, the combination of both a similarity and a dissimilarity measure, as well as the sparsity regularization.


\section{Experimental Results}
\label{sec:exps}
\noindent\textbf{Datasets:}
\begin{figure*}[t]
\centering
\begin{subfigure}[b]{0.32\textwidth}
        \includegraphics[width=\textwidth,height=8em]{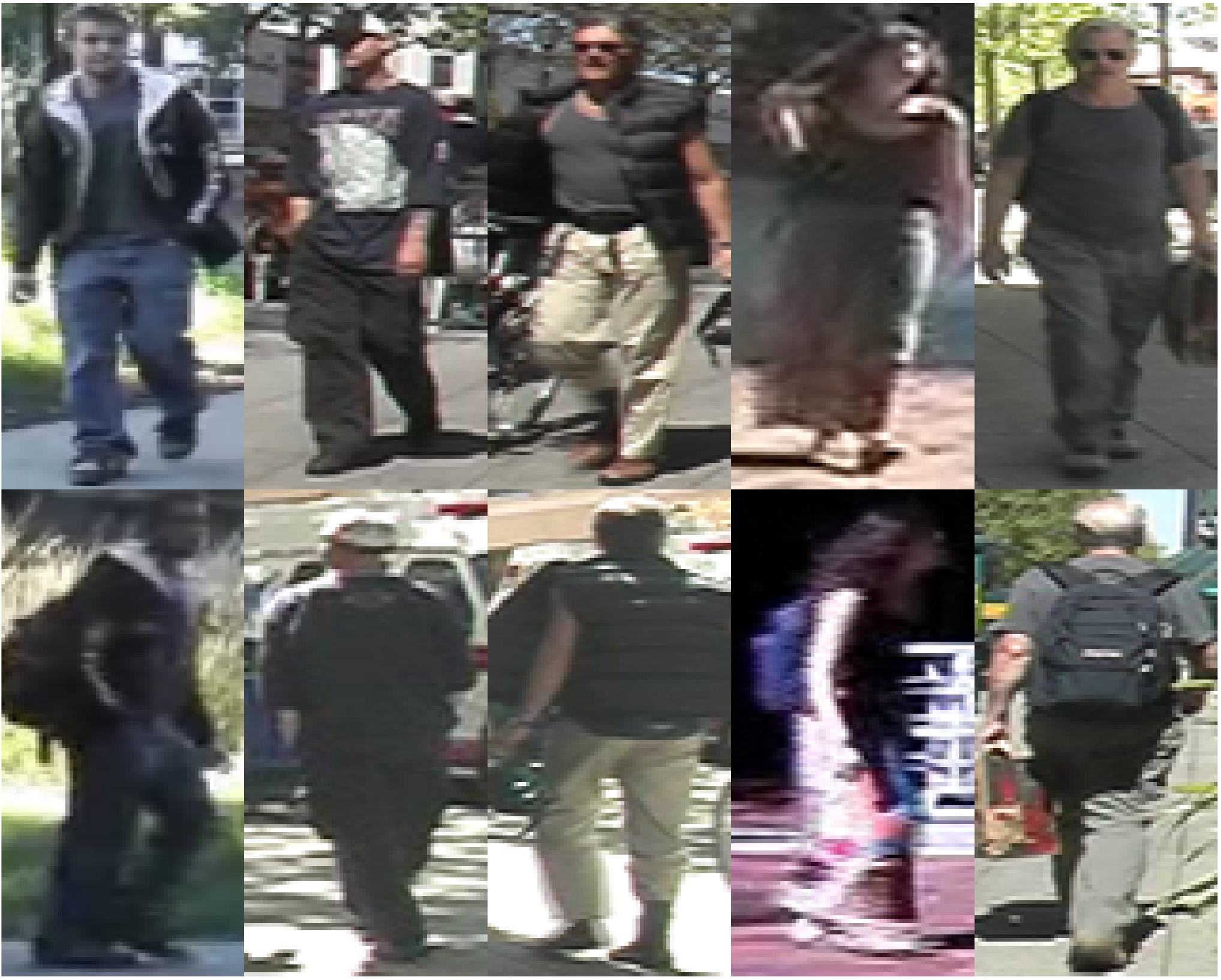}
        \caption{}
        \label{fig:viper_strip}
    \end{subfigure}
    \hfill
	\begin{subfigure}[b]{0.32\textwidth}
        \includegraphics[width=\textwidth,height=8em]{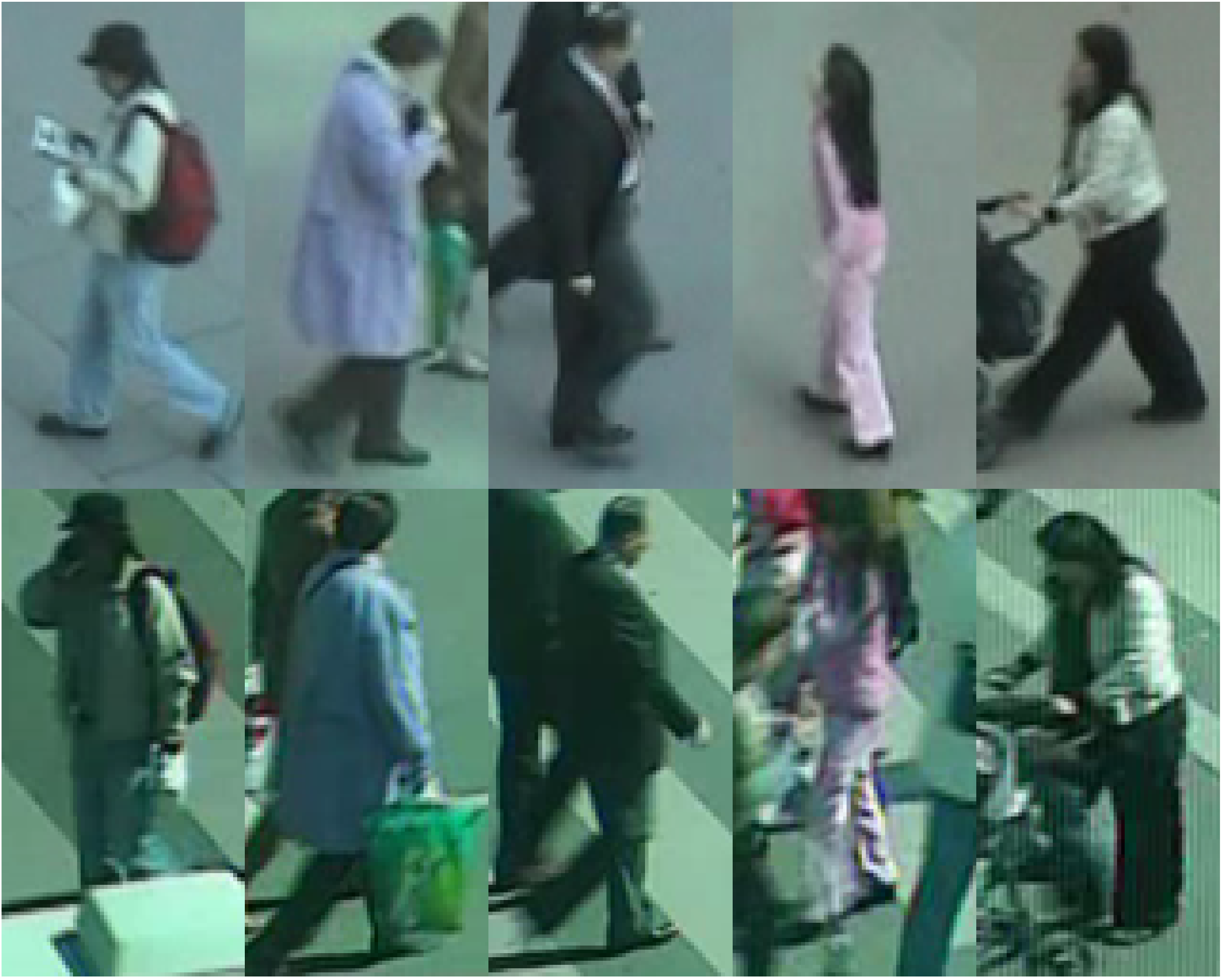}
        \caption{}
        \label{fig:prid450s_strip}
    \end{subfigure}
    \hfill
    \begin{subfigure}[b]{0.32\textwidth}
        \includegraphics[width=\textwidth,height=8em]{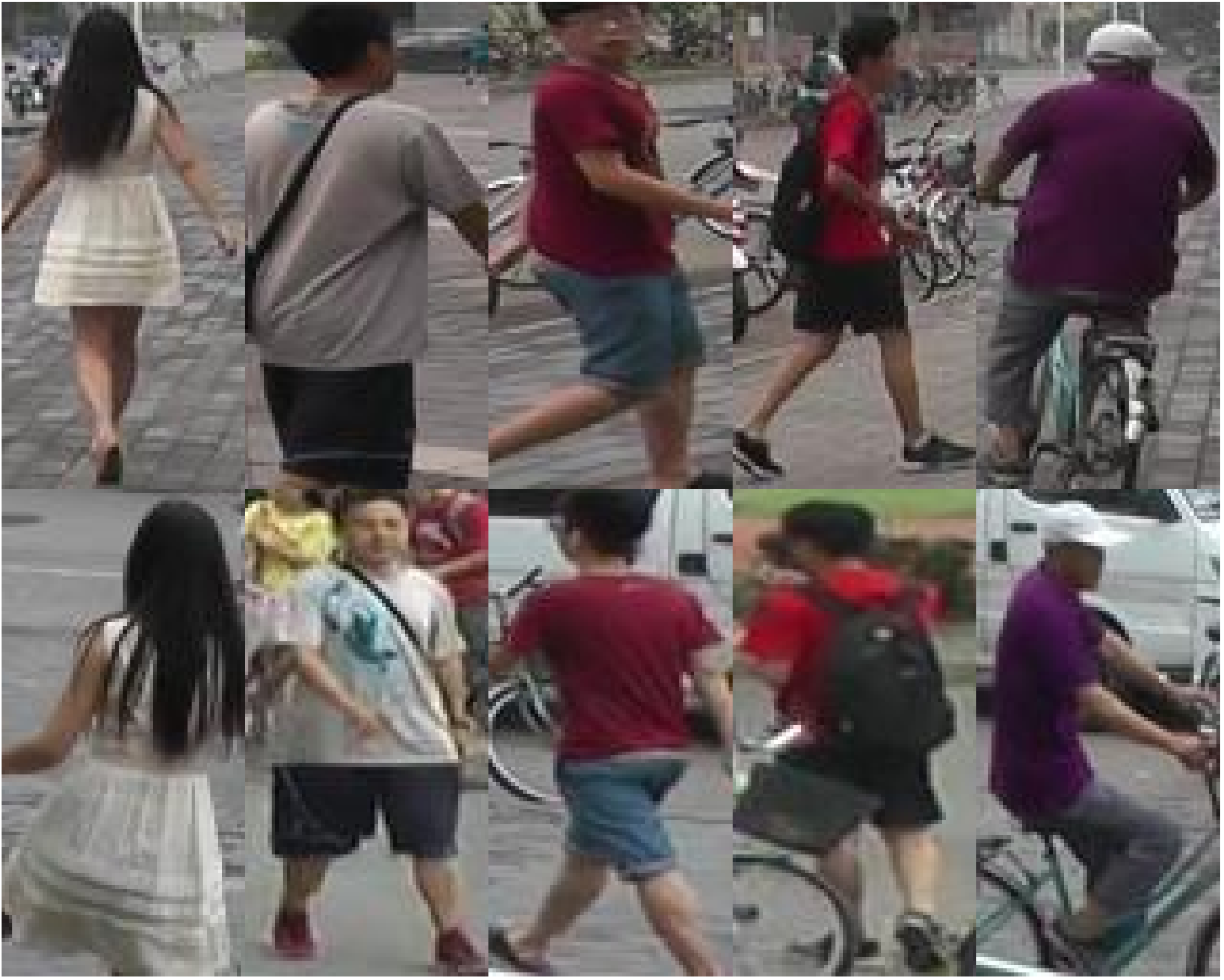}
        \caption{}
        \label{fig:market1501_strip}
    \end{subfigure}
\caption{
15 image pairs from the \subref{fig:viper_strip}~VIPeR, \subref{fig:prid450s_strip}~PRID450S and \subref{fig:market1501_strip}~Market1501 datasets.
Columns correspond to different persons, rows to different cameras.}
\label{fig:datasets}
\vspaceafterfigure
\end{figure*}
We evaluated our approach on three publicly available benchmark datasets\footnote{See supplementary for additional results on the 3DPeS and CUHK03 datasets.}, namely VIPeR~\cite{Gray2007a}, PRID450S~\cite{Roth2014a}, and Market1501~\cite{Zheng2015} (see \figurename~\ref{fig:datasets} for few sample images).
Following the literature, we run 10 trials on the VIPeR and PRID450S dataset, while we use the available partitions for Market 1501.
We report on the average performance using the Cumulative Matching Characteristic (CMC).
We refer to our method as Temporal Model Adaptation ($\myalgoname$).

\textbf{VIPeR}~\cite{Gray2007a} is considered one of the most challenging datasets.
It contains 1,264 images of 632 persons viewed by two cameras.
Most image pairs have viewpoint changes larger than 90\textdegree.
Following the general protocol, we split the dataset into a training and a test set each including 316 persons.

\textbf{PRID450S}~\cite{Roth2014a} is a more recent dataset containing 450 persons viewed by two disjoint cameras with viewpoint changes, background interference and partial occlusion.
As performed in literature~\cite{Yang2014,Shen2015}, we partitioned the dataset into a training and a test set each containing 225 individuals.


\textbf{Market1501}~\cite{Zheng2015} is the largest currently available person re-identification dataset.
It contains 32,668 images of 1,501 persons taken from 6 disjoint cameras.
Multiple images of a same person have been obtained by means of a state-of-the-art detector, thus providing a realistic setup.
To run the experiments, we used the available code\footnote{\url{http://www.liangzheng.com.cn}} to get the same BoW feature representation as well as the same train/test partitions containing 750 and 751 person identities each.


\noindent\textbf{Implementation:}
\label{sec:implementation_details}
To model person appearance we adopted the Local Maximal Occurrence (LOMO) representation~\cite{Liao2015}.
We selected $\kernellowrankpar = 0.001$, $\metriclowrankpar = 0.001$, $\learningrate = 1$, and $\admmpenalty=1$ by performing 5-fold cross validation on $\{1, 0.5, 0.1, 0.05, 0.01, 0.001\}$.
The temporal model adaptation followed the common batch framework used in active learning~\cite{Settles2012}.
It partitioned each training set into 4 disjoint \textit{batches}.
Due to the adopted randomization procedure, each batch contains approximately $z=(|\probeset|/4)\times(|\galleryset|/4)$ pairs\footnote{The percentage of labeled pairs is computed with respect to $n$.}.
We have used the first batch to train the initial model with $T=2z$, $S=200$, and no further stopping criteria.
The remaining ones have been used for the batch-incremental updates with $\hat{T}=2z$ and $\hat{S}=150$ (in the following, the subscript of $\myalgoname$ indicates the number of model updates that are achieved for every probe in each batch).
Finally, to select the relevant gallery images in each batch we have set $\epsilon=0.1$
(see Table~\ref{tab:dsepsilon}).

\subsection{State-of-the-art Comparisons}
In the following we compare the results of our approach with existing methods.
In addition to the incremental performance, we also provide our results when no model adaptation is exploited and all the training data is included in one single batch ($\myalgoname_0$).

\begin{table}[t]
\vspace*{-1em} 
\scriptsize
	\centering
	 \caption{Comparison with state-of-the-art methods on the VIPeR dataset.
 Best results for each rank are in boldface font.}
   	\label{tab:VIPER_comparison}
	\begin{tabulary}{1\linewidth}{L{8em}||C{4em}|C{4em}|C{4em}|C{4em}||C{6em}||L{10em}}
	\toprule
    Rank $\rightarrow$   & 1  & 10 & 20 & 50  & Labeled [\%] & Reference \\
	\midrule
	$\myalgoname_4$+LADF	& \textbf{48.19} & \textbf{87.65} & 93.54 & 98.41 & 20.32+100 & Proposed + \cite{Li2013b}  \\
	$\myalgoname_0$		& 43.83 & 83.86 & 91.45 & 97.47 & 100 & Proposed \\
	LMF+LADF	& 43.29 & 85.13	& \textbf{94.12} & \NA	& 100 & CVPR 2014~\cite{Zhao2014}+\cite{Li2013b}\\
	$\myalgoname_4$			& 41.46 & 82.65 & 92.46 & \textbf{99.65} & 20.32 & Proposed \\
    MLAPG 		& 40.73 & 82.34 & 92.37 & \NA	& 100 & ICCV 2015~\cite{Liao2015a}\\
	XQDA		& 40.00	& 80.51	& 91.08	& \NA	& 100 & CVPR 2015~\cite{Liao2015}\\
	$\myalgoname_3$			& 37.97 & 75.00 & 87.66 & 96.52 & 12.56 & Proposed \\
	SCNCDFinal		& 37.80 & 81.20 & 90.40 & 97.0	& 100 & ECCV 2014~\cite{Yang2014}\\
	PKFM		& 36.8	& 83.7	& 91.7	& 97.8	& 100 & CVPR 2015~\cite{Chen2015} \\
	$\myalgoname_2$			& 36.08 & 71.84 & 81.96 & 94.62 & 6.78 & Proposed \\
	$\myalgoname_1$			& 35.13 & 69.94 & 81.01 & 93.35 & 4.91 & Proposed \\
	QALF		& 30.17	& 62.44	& 73.81	& \NA	& 100 & CVPR 2015~\cite{Zheng2015}\\
	ISR			& 27.43	& 61.06	& 72.92	& 86.69	& 100 & TPAMI 2015~\cite{Lisanti2014} \\
	WFS			& 25.81	& 69.56	& 83.67	& 95.12	& 100 & TPAMI 2015~\cite{Martinel2015a} \\
	KISSME		& 19.60 & 62.20 & 77.00 & 91.80 & 100 & CVPR 2012~\cite{Kostinger2012} \\
	\bottomrule	
	\end{tabulary}
	\vspaceaftertable
\end{table}

\textbf{VIPeR:}
Results in Table~\ref{tab:VIPER_comparison} show that our approach has better performance than recent solutions even in the case only about $5\%$ of the data is used.
This result indicates that, partially due to the feature representation (see results of KISSME in Table~\ref{tab:simdis_comparison}), our approach produces a robust solution to viewpoint variations.
Incremental updates bring $\myalgoname_{4}$ to be the second best.
In such a case, only LMF+LADF performs better.
However, such an approach is a combination of two methods, which, as shown in~\cite{Xiong2014}, generally improves the performance.
Indeed, a rank 1 recognition rate of $48.19\%$ is achieved by summing $\myalgoname_0$ and LADF scores.
If the same batches as $\myalgoname_{1-4}$ are considered to train LADF, the fused rank 1 performances are of $35.6\%$, $37.9\%$, $40.8\%$ and $43.4\%$, respectively --which represent an average improvement of $11\%$ over standalone LADF.

Finally, results obtained with $\myalgoname_0$ show that the best rank 1 is achieved, but performance on higher ranks is slightly worse than the one obtained using incremental updates ($\myalgoname_4$).
Hence, using all the available data requires additional manual labor and might also drive to decreasing performance.
This strengthens our contribution showing that, by identifying the most informative samples to train with, better results can be achieved with reduced human effort.

	\begin{table}[t]
\vspace*{-1em} 
\scriptsize
	\centering
	 \caption{Comparison with state-of-the-art methods on the PRID 450S dataset.
 Best results for each rank are in boldface font.
}
   	\label{tab:PRID450S_comparison}
	\begin{tabulary}{1\linewidth}{L{8em}||C{3.5em}|C{3.5em}|C{3.5em}|C{3.5em}|C{3.5em}||C{6em}||L{8em}}
	\toprule
    Rank $\rightarrow$   & 1  & 5 & 10 & 20 &  50  & Labeled[\%] & Reference \\
    \midrule
    $\myalgoname_0$ & \textbf{54.22} & 73.78 & 83.11 & 90.22 & 97.33 & 100 & Proposed \\
	$\myalgoname_4$ & 52.89 & \textbf{76.00} & \textbf{85.78} & \textbf{93.33}  & \textbf{97.78} & 14.25 & Proposed \\
    $\myalgoname_3$	& 50.22 & 75.56 & 85.33 & 92.89 & 97.64 & 10.18 & Proposed \\
    $\myalgoname_2$	& 48.89 & 75.33 & 84.01 & 91.1 & 97.33 & 8.64 & Proposed \\
    $\myalgoname_1$	& 45.33 & 72.00 & 83.11 & 89.78 & 96.02 & 6.42 & Proposed \\    		
	CSL        	& 44.4 	& 71.6	& 82.2 	& 89.8 	& 96.0 	& 100 	& ICCV2015~\cite{Shen2015}\\ 
	SCNCDFinal 	& 41.6 	& 48.9 	& 79.4 	& 87.8 	& 95.4 	& 100 	& ECCV2014~\cite{Yang2014}\\ 
	SCNCD      	& 41.5 	& 66.6 	& 75.9 	& 84.4 	& 92.4 	& 100 	& ECCV2014~\cite{Yang2014} \\ 
	KISSME     	& 33   	&   \NA   	& 71   	& 79   	&   90  & 100 	& CVPR2012~\cite{Kostinger2012} \\ 

	\bottomrule
	\end{tabulary}
	\vspaceaftertable
\end{table}
\textbf{PRID450S:}
In Table~\ref{tab:PRID450S_comparison} we report on the performance comparisons between existing method and our approach on the PRID450S dataset.
Results show that our solution outperforms the methods used for comparisons regardless of the amount of data used for training.
In particular, using only $14.25\%$ of the data an $8\%$ improvement with respect to the best existing approach is obtained at rank 1.
By training only with the initially available data (\ie, $\myalgoname_1$), our solution outperforms SCNCDFinal~\cite{Yang2014}, which, on the VIPeR dataset, had better performance (until the $3^{rd}$ batch update).
This may suggest that our approach is robust to background clutter and occlusions which PRID450S suffer from.

\begin{table}[t]
\vspace*{-1em} 
\scriptsize
	\centering
	 \caption{Rank 1 and mAP performance comparison with existing methods on the Market 1501 dataset.
	 Best result is in boldface font.}
   	\label{tab:market1501_comparison}
   	\begin{adjustwidth}{-0.5em}{}
   	\begin{tabulary}{1\linewidth}{C{5.5em}||C{3.5em}|C{4.2em}|C{4em}|C{4.2em}|C{3.5em}|C{3.5em}|C{3.5em}|C{3.5em}||C{4.5em}}
	\toprule
	Method & BoW \cite{Zheng2015} & LMNN BoW \cite{Zheng2015} & ITML BoW\cite{Zheng2015} & KISSME BoW \cite{Zheng2015} &  $\myalgoname_1$ BoW &  $\myalgoname_2$ BoW &  $\myalgoname_3$ BoW & $\myalgoname_4$ BoW & $\myalgoname_4$ LOMO  \\ 
	 \midrule
	Rank 1 & 42.64 & 38.91 & 27.08 & 43.03 & 28.77 & 34.68 & 39.81 & 44.74 & \textbf{47.92} \\
	mAP & 19.47 & 17.34 & 8.13 & 19.98 & 8.69 & 14.56 & 17.89 & 20.92 & \textbf{22.31} \\
	\midrule   
	Labeled [\%] & 100 & 100 & 100 & 100 & 5.23 & 8.71 & 12.09  & 14.36 & 13.58 \\	
	\bottomrule	
	\end{tabulary}
	\end{adjustwidth}
	\vspaceaftertable
\end{table}
\textbf{Market1501:}
Comparisons of our approach with existing methods on the Market 1501 dataset are shown in Table~\ref{tab:market1501_comparison}.
The obtained performance are consistent with the ones achieved on the VIPeR and PRID450S datasets.
Our approach has significantly better performance than methods used for comparisons even by using $5.23\%$ of labeled data.
Incremental updates bring in relevant improvements and with $\myalgoname_4$ we achieve the best rank 1 recognition rate,~\ie,$44.74\%$.
Using the LOMO feature representation instead of the BoW one provided by~\cite{Zheng2015}, about a $3\%$ rank 1 performance gain is obtained.
Results on such dataset demonstrate that our approach can scale to a real scenario and achieve competitive performance with significantly less manual labor.
The reason for the improved performance with much less training data is because our method identifies the most discriminating examples to train with, and does not waste labeling effort on those that will add little or no value to the re-identification accuracy.

\subsection{Influence of the Temporal Model Adaptation Components}
To better understand the achieved performance, we have run additional experiments by separately considering the similarity-dissimilarity metric learning approach and the probe relevant set selection method.

\noindent\textbf{Similarity-Dissimilarity Metric:}
\label{sec:sub:simdisperf}
In the following, we first analyze the contribution of the similarity and the dissimilarity components.
Then, we compare our performance with existing methods using the same LOMO representation.

\begin{figure*}[t]
\centering
	\begin{subfigure}[b]{0.243\textwidth}
        \includegraphics[width=\textwidth]{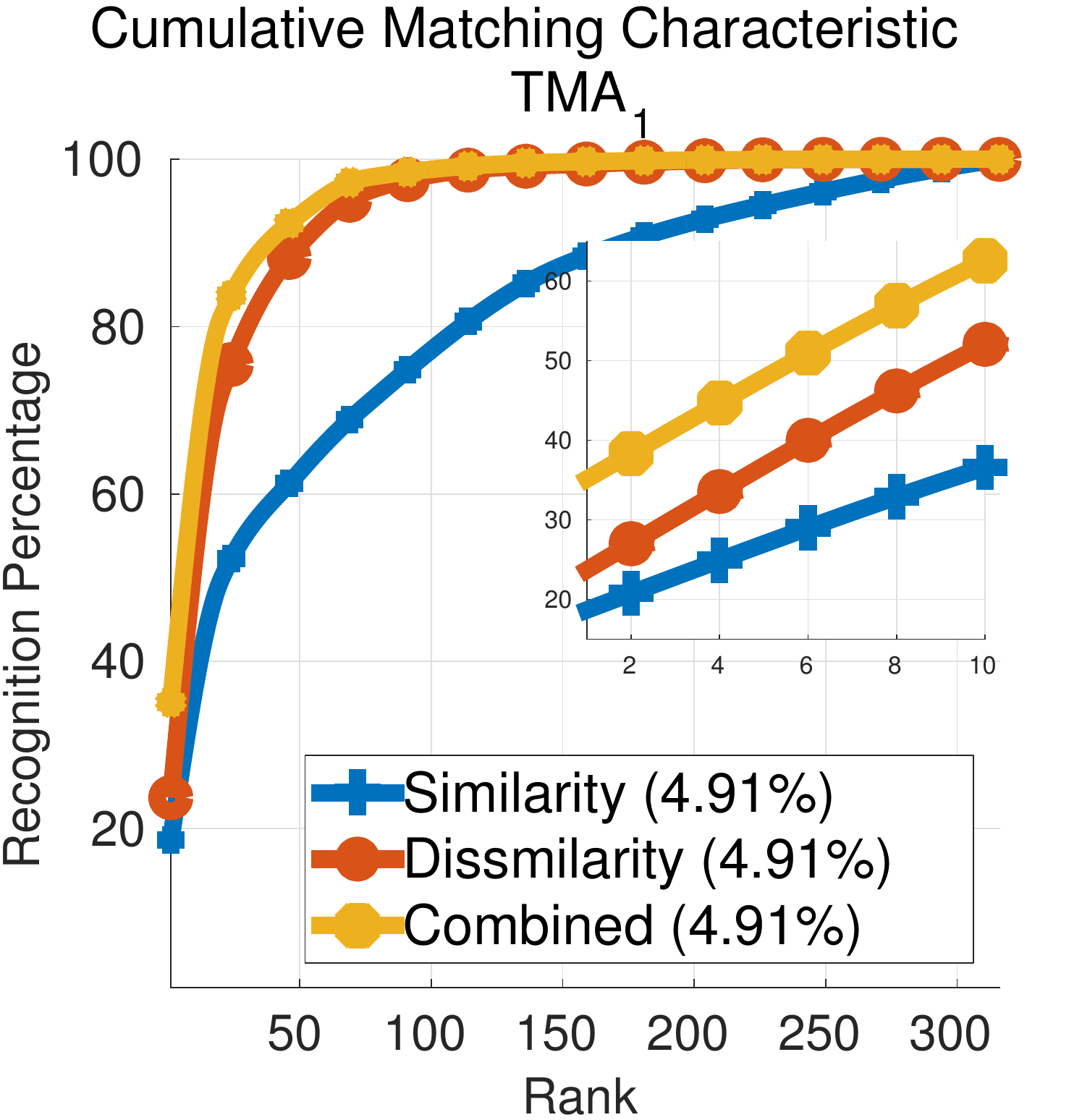}
        \caption{}
        \label{fig:viper_tybe_b1}
    \end{subfigure}
	\begin{subfigure}[b]{0.243\textwidth}
        \includegraphics[width=\textwidth]{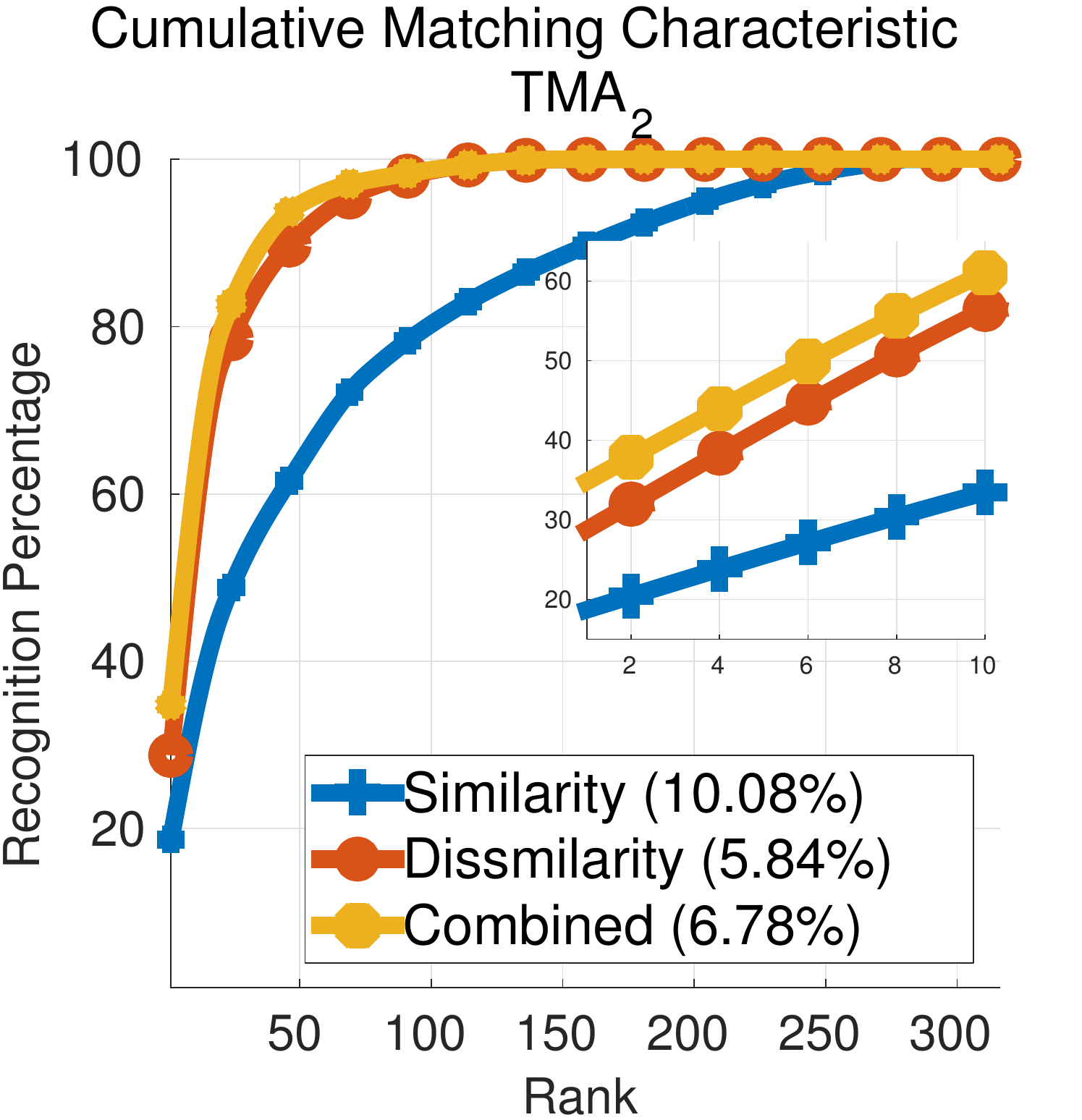}
        \caption{}
        \label{fig:viper_tybe_b2}
    \end{subfigure}
    \begin{subfigure}[b]{0.243\textwidth}
        \includegraphics[width=\textwidth]{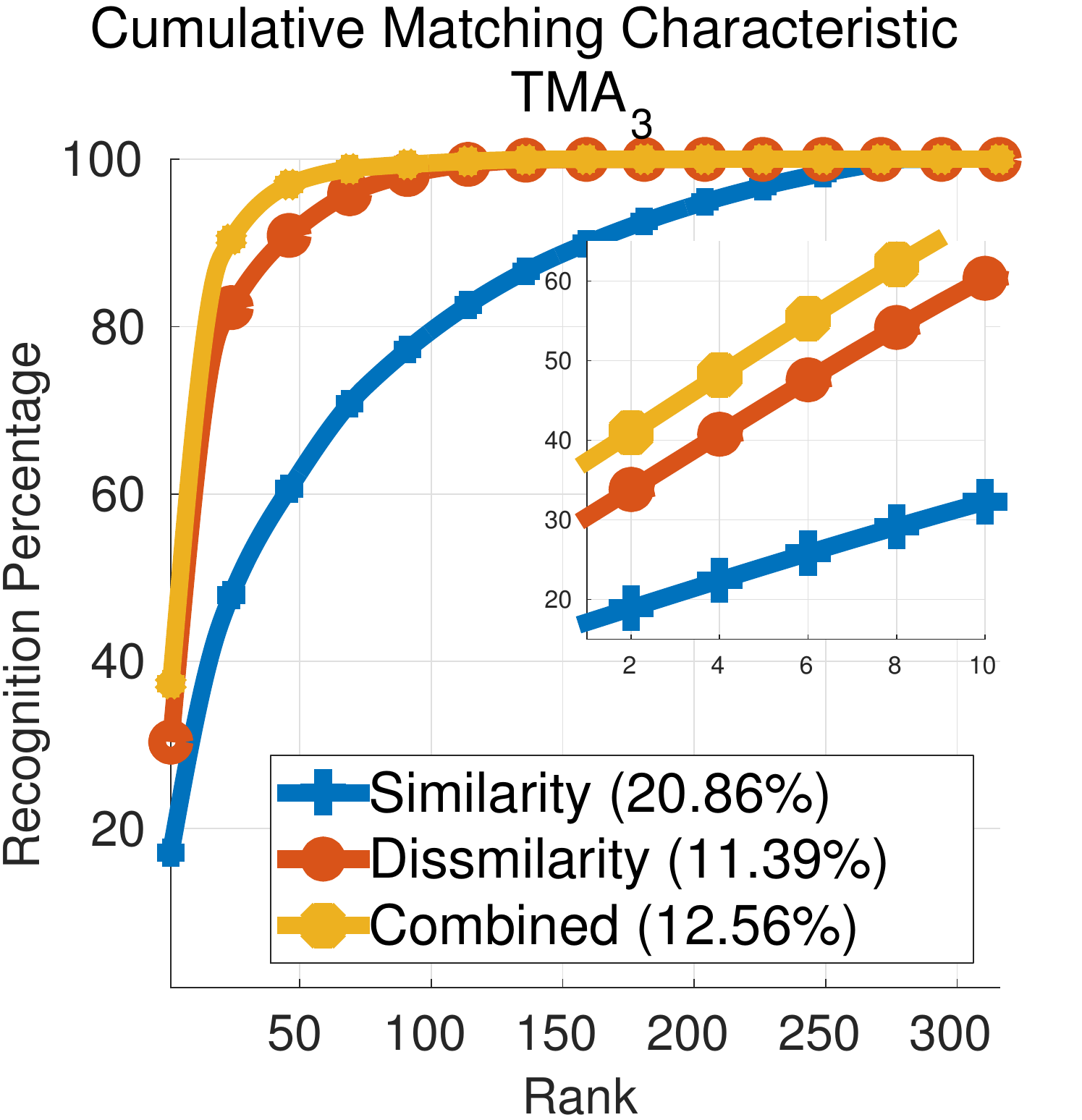}
        \caption{}
        \label{fig:viper_tybe_b3}
    \end{subfigure}
    \begin{subfigure}[b]{0.243\textwidth}
        \includegraphics[width=\textwidth]{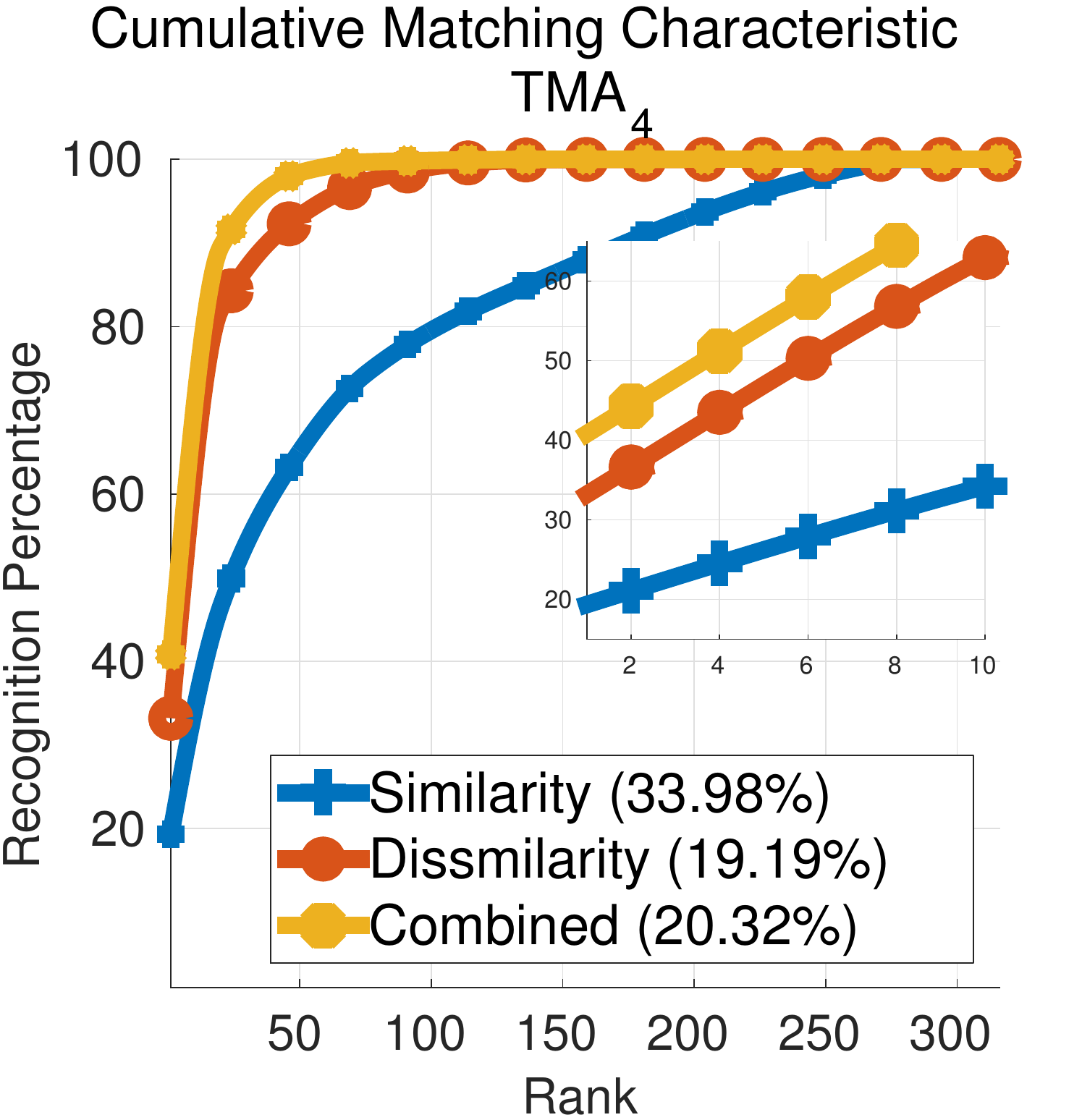}
        \caption{}
        \label{fig:viper_tybe_b4}
    \end{subfigure}
\caption{Comparison of the similarity-dissimilarity learning components. 
\subref{fig:viper_tybe_b1}--\subref{fig:viper_tybe_b4} show the results on the VIPeR dataset computed using incremental batch updates.
For each curve, the percentage of manually labeled samples is indicated in parenthesis.
The inside picture show the results for rank range 1--10.}
\label{fig:simdisperf}
\vspaceafterfigure
\end{figure*}

\textbf{\textit{Contribution of the components:}}
In~\figurename\ref{fig:simdisperf}, we report on the results obtained using either the learned similarity, the learned dissimilarity or both.
Results show that most of the performance contribution is provided by the dissimilarity.
The similarity has significantly lower performance and calls for more labeled pairs.
This is due to the fact that the majority of the edges of the corresponding graph have weak weights, thus causing the maximization procedure to select more samples before the stop condition.
Enforcing agreement on a specific pair by jointly optimizing the similarity and the dissimilarity measure results in the best performances.
With respect to the dissimilarity approach, this yields negligible increase of manual labor and improved results ($7\%$ at rank 1).

\begin{table*}[t]
\vspace{-1em}	
\scriptsize
\centering
\caption{Comparison with metric learning approaches on the VIPeR dataset. Results obtained using truncated projections (100 dimensions) are given for three representative ranks. Last row shows the percentage of manually labeled samples. Best results for each rank are in bold. Most of the results are from~\cite{Liao2015a}.}
\label{tab:simdis_comparison}
\begin{adjustwidth}{-1.5em}{}
\begin{tabular}{c||c|c|c|c|c|c|c|c|c|c|c|c}
\toprule
Rank $\downarrow$ & MLAPG & XQDA & KISSME & LMNN & LADF & ITML & LDML & PRDC & $\myalgoname_1$ & $\myalgoname_2$ & $\myalgoname_3$ & $\myalgoname_4$ \\
\midrule
 1 & 39.21 & 38.23 &  33.54 & 28.42 & 27.63 &  19.02& 13.99 & 12.15 & 32.28 & 34.81  & 36.07  & \textbf{39.88} \\
 10 & \textbf{81.42} & 81.14 & 79.30 & 72.31 & 75.47 & 52.31 & 38.64 & 35.82 & 69.62 & 73.10  & 76.27  & 81.33 \\
 20 & \textbf{92.50}  & 92.18 & 90.47 & 85.32 & 88.29 & 67.34 & 48.73 & 48.26 & 81.33 & 58.79  & 90.19  & 91.46 \\
\midrule
Labeled [\%] & 100 & 100 &  100 & 100 & 100 & 100 & 100 & 100 & 4.91 & 6.91  & 11.48  & 15.77  \\
\bottomrule
\end{tabular}
\end{adjustwidth}
\end{table*}

\textbf{\textit{Comparison with existing methods:}}
In Table~\ref{tab:simdis_comparison}, we report on the comparison of our similarity\hyp{}dissimilarity approach with general state-of-the-art metric learning approaches, namely ITML~\cite{Davis2007}, LMNN~\cite{Weinberger2009}, LDML~\cite{Guillaumin2009}, and re-identification tied ones namely, PRDC\cite{Zheng2012a}, KISSME~\cite{Kostinger2012}, LADF~\cite{Li2013b}, XQDA~\cite{Liao2015}, and MLAPG~\cite{Liao2015a}.
To provide a fair comparison, we used the same settings in~\cite{Liao2015a}.
Precisely, the 100 principal components found by PCA have been exploited to train LMNN, ITML, KISSME, and LADF.
Since other methods,~\ie, XQDA, PRDC, LDML, MLAPG and $\myalgoname$, are able to discover the discriminative features, we used all the principal components.
For a fair comparison, projection learned by XQDA, MLAPG and $\myalgoname$ were truncated to 100 dimensions.

Results in Table~\ref{tab:simdis_comparison} show that our approach, trained with only $4.91\%$ of the available data, has the $4^{th}$ best rank 1 result.
As shown in~\figurename~\ref{fig:simdisperf}, such a successful result is due to the competition between the similarity and the dissimilarity approaches.
Performing incremental updates yields significant improvements and, after the $4^{th}$ update is completed, the best rank 1 recognition rate is achieved.
At higher ranks, $\myalgoname$ performs on par with other methods but with substantially less labeled pairs (\ie, $15.77\%$ of all possible annotations).

\textbf{\textit{Discussion:}}
Results have demonstrated that, while the dissimilarity metric has more impact on the performance, by enforcing competition with the similarity measure better results can be obtained.
Additional evaluations showed that by removing the $\ell_{2,1}$ norms the degradation is of 3\%.
Comparisons with existing approaches have shown that, under the same conditions, our approach achieves good results using only $1/6$ of the data.
Incremental updates produce considerable improvements with a significantly reduced human effort.
This substantiates the benefits of the proposed similarity-dissimilarity learning approach and demonstrate the feasibility of temporal model adaptation for the task. 

\noindent\textbf{Probe Relevant Set Selection:}
\label{sec:sub:dsinfluence}
In the following, we provide an analysis of the graph-based solution to identify the most informative gallery persons.
We report on the effects of the $\epsilon$ parameter, then we compare with three approaches.

\begin{table*}[t]
\vspace*{-1em} 
\scriptsize
\centering
\caption{Analysis of the $\epsilon$ parameter used to obtain the probe relevant set. Each entry in the table shows the rank 1 performance as well as the percentage of labeled data (in brackets). Best results for each rank are in bold.}
\label{tab:dsepsilon}
\begin{tabular}{C{4em}||C{7.5em}|C{7.5em}|C{7.5em}|C{7.5em}|C{7.5em}}
\toprule
$\epsilon \rightarrow$ & 0.5 & 0.3 & 0.1 & 0.05 & 0.01 \\
\midrule
$\myalgoname_1$  & 35.13 (4.91) & 35.13 (4.91) & 35.13 (4.91) & 35.13 (4.91) & 35.13 (4.91)  \\
$\myalgoname_2$  & \textbf{36.08} (7.87) & 35.76 (7.26) & \textbf{36.08} (6.78) & 34.49 (6.62) & 34.49 (6.20)  \\
$\myalgoname_3$  & 37.03 (11.36) & 36.23 (10.59) & \textbf{37.97} (12.56) & 36.71 (8.97) & 34.81 (7.95) \\
$\myalgoname_4$  & 38.61 (14.74) & 38.92 (13.50) & \textbf{41.46} (20.32) & 39.87 (11.49) & 37.97 (9.81) \\
\bottomrule
\end{tabular}
\vspaceaftertable
\end{table*}

\textbf{\textit{Influence of $\epsilon$:}}
To verify the influence of the $\epsilon$ parameter, we have computed the results in Table~\ref{tab:dsepsilon}.
These show that, large values of $\epsilon$ produce coarse under-segmented sets, hence identify a large number of relevant pairs to label.
Small values of $\epsilon$,~\eg, $0.01$, produce over segmented-graphs, hence small dominant sets.
Indeed, after the $4^{th}$ update, less than $10\%$ of all the available pairs has been used for training.
This results in achieving similar performance improvements, but with a different manual effort.
The reason behind this is that, in the former case, the probe relevant sets contain additional persons which are not ``similar'' to the probe and any other gallery person.
This causes the model to be updated with uninformative pairs which weaken its discriminative power.
In the latter, too few informative pairs are found and the model overfits such samples.

\begin{figure*}[t]
\centering
	\begin{subfigure}[b]{0.31\textwidth}
        \includegraphics[width=\textwidth]{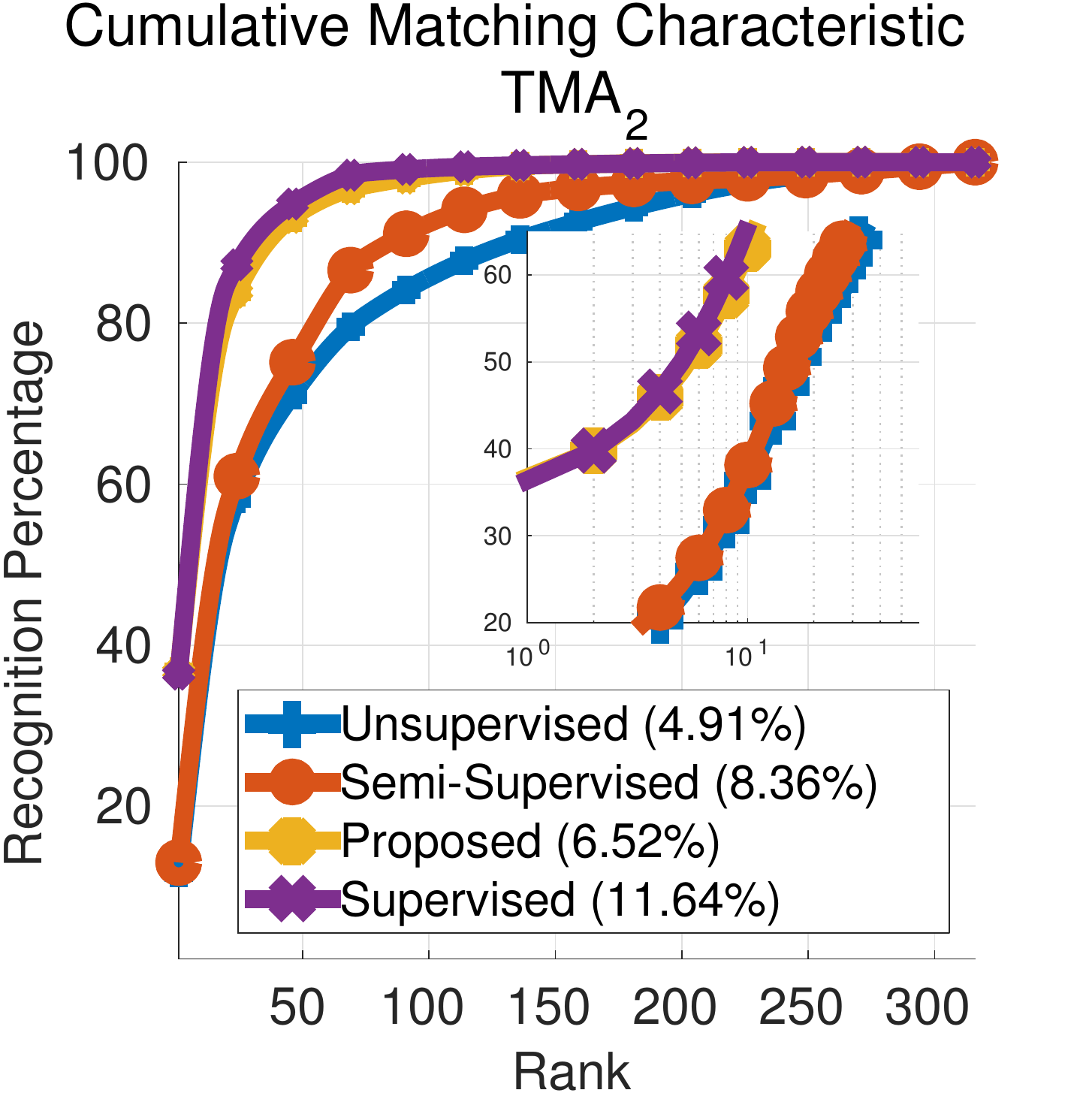}
        \caption{}
        \label{fig:viper_ds_b2}
    \end{subfigure}
    \hfill
    \begin{subfigure}[b]{0.31\textwidth}
        \includegraphics[width=\textwidth]{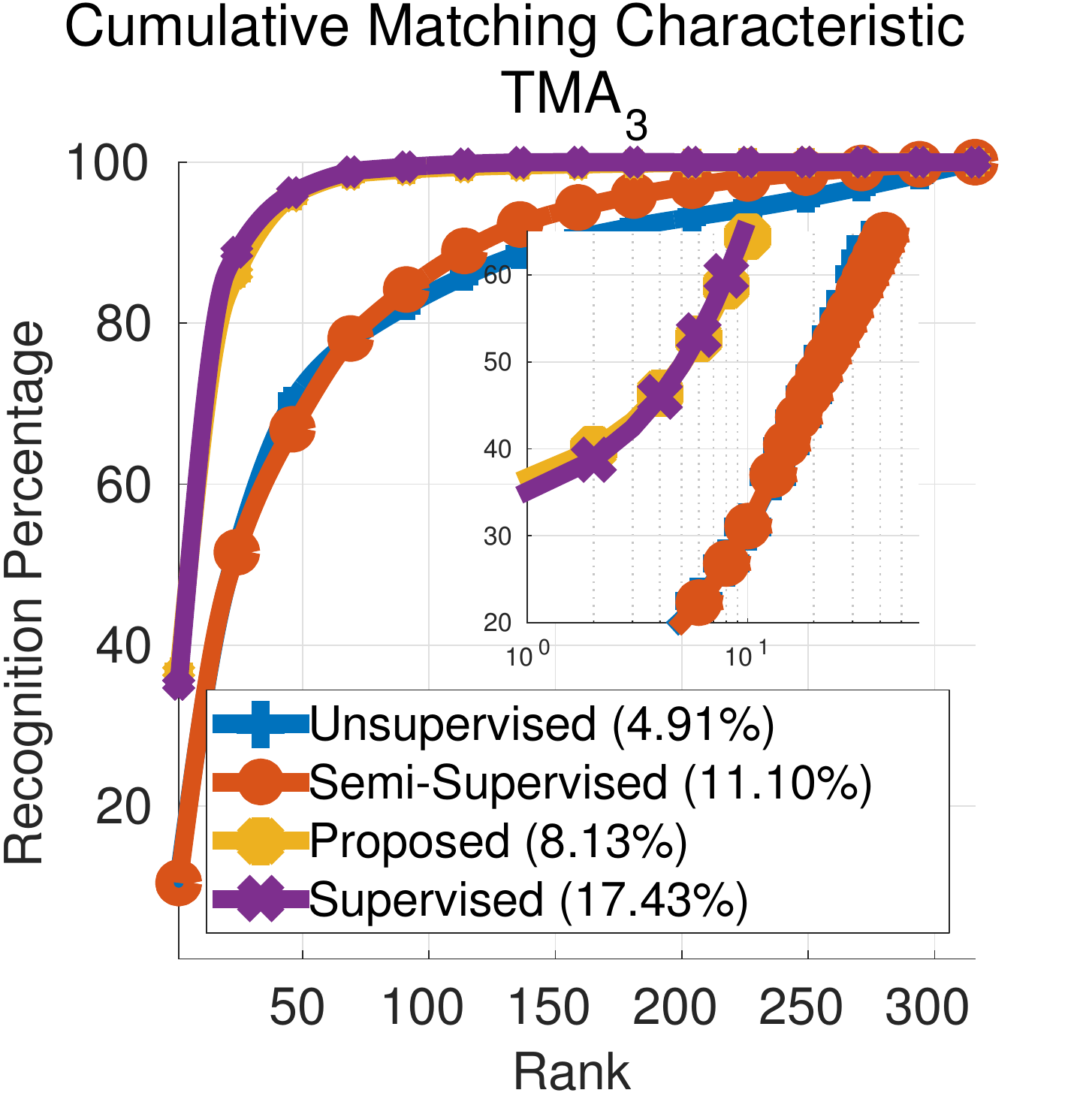}
        \caption{}
        \label{fig:viper_ds_b3}
    \end{subfigure}
    \hfill
    \begin{subfigure}[b]{0.31\textwidth}
        \includegraphics[width=\textwidth]{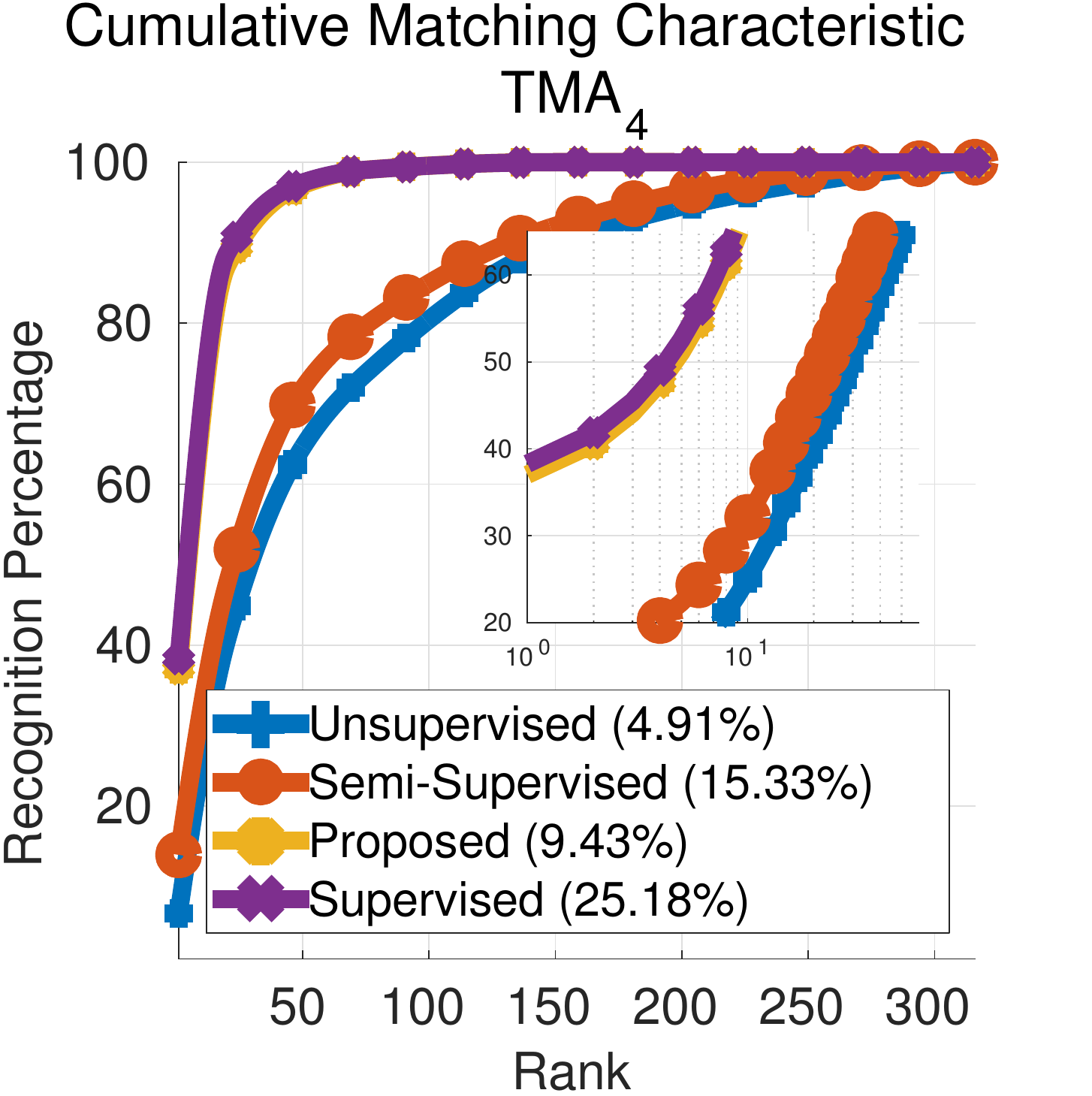}
        \caption{}
        \label{fig:viper_ds_b4}
    \end{subfigure}
\caption{Re-Identification performance on the VIPeR dataset computed using four different probe relevant set selection criteria.
~\subref{fig:viper_ds_b2}-\subref{fig:viper_ds_b4} show the performances achieved using the $2^{nd}$-$4^{th}$ batch incremental updates.
The percentage of manually labeled samples is given within parenthesis.
The inside picture show the results on a log-scale reduced rank range,~\ie,1-50.}
\label{fig:ds_criteria}
\vspaceafterfigure
\end{figure*}
\textbf{\textit{Selection Criteria Comparison:}}
In~\figurename\ref{fig:ds_criteria}, we compare our probe relevant set selection approach with three different criteria.
Before exploiting such criteria, we applied Platt scaling~\cite{Platt1999} to the obtained scores to get the probability of each probe-gallery pair being positive.
\begin{enumerate}[label=\roman{*}), topsep=1pt,itemsep=-1ex,partopsep=1ex,parsep=1ex]
\item \textit{Unsupervised}: Each pair having probability less than $0.5$ has been assigned the negative label, remaining ones have been assigned the positive label.
\item \textit{Semi-Supervised}: Top and bottom 20 ranked pairs have been labeled as positive or the negative, respectively.
Remaining pairs have been human labeled.
\item \textit{Supervised}: Every pair has been human labeled.
\end{enumerate}
Results show that using the unsupervised or the semi-supervised criteria, the performance obtained with incremental updates tends to decrease.
This behavior is due to the fact that, right after the first update, the produced scores induce very small or very large probabilities.
This yields zero manual labor, but, as a consequence, the model is updated with a large portion of mislabeled samples.
Using our solution, performance reaches the ones obtained using a fully-supervised approach.
In particular, with the $4^{th}$ batch update our approach yields the highest rank 1 recognition rate ($41.46\%$ vs $39.87\%$) with $5\%$ less manual labor.
Additional experiments considering the human mislabeling error $C\in\{5,\ldots,95\}\%$ show that the model update is effective when $C\leq15\%$.

\textbf{\textit{Discussion:}}
In this section, we have shown that our approach is moderately sensible to the selection of $\epsilon$, which to some extent, controls the human effort.
In addition, it performs better than a fully supervised approach in which all the samples are manually labeled.
This demonstrates that the proposed approach identifies the most informative pairs that should be used to update the model.

\subsection{Computational Complexity}
\vspace{-2.5em} 
\begin{table*}[h!t]
\vspace*{-1em} 
\scriptsize
\centering
\caption{Comparison between deterministic ADMM and our stochastic solution.
VIPeR result computed by running MATLAB code on an Intel Xeon 2.6GHz. Complexity is computed for the parameters updates which differs from the two solutions.}.
\label{tab:admm_scasadmm}
\begin{tabulary}{1\linewidth}{L{11em}||C{9em}|C{11em}|C{11em}}
\toprule
Method $\downarrow$ & $\myalgoname_1$ - Rank 1 & Per-Epoch Complexity & Training Time [s]  \\
\midrule
Deterministic ADMM		& 34.84 & $\mathcal{O}\left(2(n2d^2+d^3)\right)$ 	& 12051.19  \\
Stochastic ADMM			& 35.15 & $\mathcal{O}\left(2(n3d^2+K3d^2)\right)$ 	& 2948.38   \\
\bottomrule
\end{tabulary}
\vspaceaftertable
\end{table*}

\vspace{-2em}
In Table~\ref{tab:admm_scasadmm}, we compare the computational performance of deterministic ADMM and our stochastic solution.
While achieving similar rank 1 performance, deterministic ADMM brings in more complexity, hence the training time is considerably higher.
In particular, while $d$ might be arbitrarily large, $n$ and $K$ are usually small (those depend on the number of samples which are manually labeled), thus our solution is more desirable in a continuous learning scenario.

Finally, notice that, while the initial training is more expensive than existing approaches, \eg, KISSME~\cite{Kostinger2012}, the proposed incremental learning solution is more effective in the long term since it does not require re-training like others.

\section{Conclusion} 
\label{sec:concl}
In this paper we have proposed a person re-identification approach based on a temporal adaptation of the learned model with human in the loop.
First, to allow temporal adaptation, we have proposed a similarity-dissimilarity metric learning approach which can be trained in an incremental fashion by means of a stochastic version of the ADMM optimization method.
Then, to update the model with the proper information, we have included the human in the loop and proposed a graph-based approach to select the most informative pairs that should be manually labeled.
Informative pairs selection has been obtained through the dominant sets graph partition technique.
Results conducted on three datasets have shown that similar or better performances than existing methods can be achieved with significantly less manual labor.


{
\small
\section*{\small Acknowledgment}
The work was partially supported by US NSF grant IIS-1316934.
}

\bibliographystyle{splncs}
\bibliography{0872}
\end{document}